\UseRawInputEncoding
\documentclass[sn-mathphys,Numbered]{sn-jnl}
\usepackage{hyperref}
\usepackage{graphicx}%
\usepackage{multirow}%
\usepackage{amsmath,amssymb,amsfonts}%
\usepackage{amsthm}%
\usepackage{mathrsfs}%
\usepackage[title]{appendix}%
\usepackage{xcolor}%
\usepackage{textcomp}%
\usepackage{manyfoot}%
\usepackage{booktabs}%
\usepackage{algorithm}%
\usepackage{listings}%
\usepackage{algorithmic}

\theoremstyle{thmstyleone}%
\theoremstyle{thmstyletwo}%

\theoremstyle{thmstylethree}%

\begin{document}

\title[Article Title]{Towards Open-World Co-Salient Object Detection with Generative Uncertainty-aware Group Selective Exchange-Masking}


\author[1]{\fnm{Yang} \sur{Wu}}
\email{ wuy98419@163.com}
\equalcont{These authors contributed equally to this work.}

\author[1]{\fnm{Shenglong} \sur{Hu}}\email{hslnuist@163.com}
\equalcont{These authors contributed equally to this work.}
\author*[1]{\fnm{Huihui} \sur{Song}}\email{songhuihui@nuist.edu.cn}
\author[1]{\fnm{Kaihua} \sur{Zhang}}\email{zhkhua@gmail.com}
\author[2]{\fnm{Bo} \sur{Liu}}\email{kfliubo@gmail.com}
\author[3]{\fnm{Dong} \sur{Liu}}\email{dongliu.hit@gmail.com}



\affil[1]{\orgdiv{B-DAT and CICAEET}, \orgname{Nanjing University of Information Science and Technology}, \orgaddress{\city{Nanjing}, \state{Jiangsu}, \country{China}}}
\affil[2]{\orgdiv{Walmart Global Tech}, \orgaddress{\city{Sunnyvale}, \state{CA, 94086}, \country{USA}}}
\affil[3]{\orgdiv{Netflix Inc, Los Gatos, CA, 95032, USA}}


\abstract{
%
The traditional definition of co-salient object detection (CoSOD) task is to segment the common salient objects in a group of relevant images. 
This definition is based on an assumption of group consensus consistency that is not always reasonable in the open-world setting, which results in robustness issue in the model when dealing with irrelevant images in the inputting image group under the open-word scenarios.
%
%
To tackle this problem, we introduce a group selective exchange-masking (GSEM) approach for enhancing the robustness of the CoSOD model. GSEM takes two groups of images as input, each containing different types of salient objects. Based on the mixed metric we designed, GSEM selects a subset of images from each group using a novel learning-based strategy, then the selected images are exchanged.
To simultaneously consider the uncertainty introduced by irrelevant images and the consensus features of the remaining relevant images in the group, we designed a latent variable generator branch and CoSOD transformer branch. The former is composed of a vector quantised-variational autoencoder to generate stochastic global variables that model uncertainty. The latter is designed to capture correlation-based local features that include group consensus.
%
%
Finally, the outputs of the two branches are merged and passed to a transformer-based decoder to generate robust predictions.
Taking into account that there are currently no benchmark datasets specifically designed for open-world scenarios, we constructed three open-world benchmark datasets, namely OWCoSal, OWCoSOD, and OWCoCA, based on existing datasets. By breaking the group-consistency assumption, these datasets provide effective simulations of real-world scenarios and can better evaluate the robustness and practicality of models.
Extensive evaluations on co-saliency detection with and without irrelevant images demonstrate the superiority of our method over a variety of state-of-the-art methods. The codes and datasets can be obtained from \url{https://github.com/wuyang98/CoSOD}.
}




\maketitle

\section{Introduction}

The goal of co-salient object detection (CoSOD) is to segment the salient objects that are common within a group of images. CoSOD is more challenging than single salient object detection (SOD) since it must separate the salient objects that appear simultaneously in several images, which are easily obscured by other distracting things that have a similar look, form, or semantics, only a handful to name~\cite{wu2023group}. The group input and the same semantic information are key factors that make CoSOD different from SOD.
%
%
Despite being highly challenging, CoSOD has garnered significant attention and made some progress in the past few years due to its powerful potential to help the downstream tasks by identifying the co-salient objects within a group of images and eliminating background and redundant content from those images, such as object tracking~\cite{yang2022co}, image retrieval~\cite{cheng2014salientshape}, co-segmentation~\cite{zhang2022deep} and semantic segmentation~\cite{zeng2019joint}, to name a few.


Since the year of 2010~\cite{jacobs2010cosaliency}, numerous existing CoSOD methods widely employ the group consensus assumption, which implies that all images contain the common salient objects. Additionally, current benchmark datasets such as CoSOD3k~\cite{fan2020taking}, CoCA~\cite{zhang2020gradient}, CoSal2015~\cite{zhang2015co}, and COCO-SEG~\cite{wang2019robust} are also organized based on this assumption, with each group of images containing prominently the objects with the same semantic category.
%
%
Considering the group consensus characteristic in modeling is a natural and established convention. For instance, early works such as~\cite{jacobs2010cosaliency,chang2011co,fu2013cluster} utilize hand-crafted features extracted by Gabor~\cite{gabor1946theory} and SIFT~\cite{lowe2004distinctive} to identify correspondences between objects across different images.
%
The recent learning-based models proposed in~\cite{zhang2019co,li2019detecting,jin2020icnet,zhang2020deep,yu2022democracy,zhang2022deep,liu2023self,li2023discriminative} are also under the group consensus assumption. 
They utilize a single set of related images as input training data to explore within-group consensus representations. Within this context, numerous innovative model design techniques have been devised to leverage the group consistency feature, including the unsupervised clustering techniques~\cite{zhang2020deep}, self-attention mechanisms~\cite{yu2022democracy}, network modulation techniques~\cite{zhang2022deep}, self-supervised manner~\cite{liu2023self} and region-to-region correlation strategy~\cite{li2023discriminative}.
%
The limitation of this assumption is partially studied in recent literature\cite{fan2021group}. In this work, not only the intra-group consistency is considered, but also the inter-group separability is modeled by the group collaborating module. Although it has achieved some results, it still has some limitations when there are noisy images exist in a single group.
%
Building upon this assumption, the previous models presume that the consistent information extracted between groups applies universally to all images.

\begin{figure*}[!t]
\begin{center}
\begin{tabular}{c}
\hspace{-0.18cm}\includegraphics[width=1\textwidth]{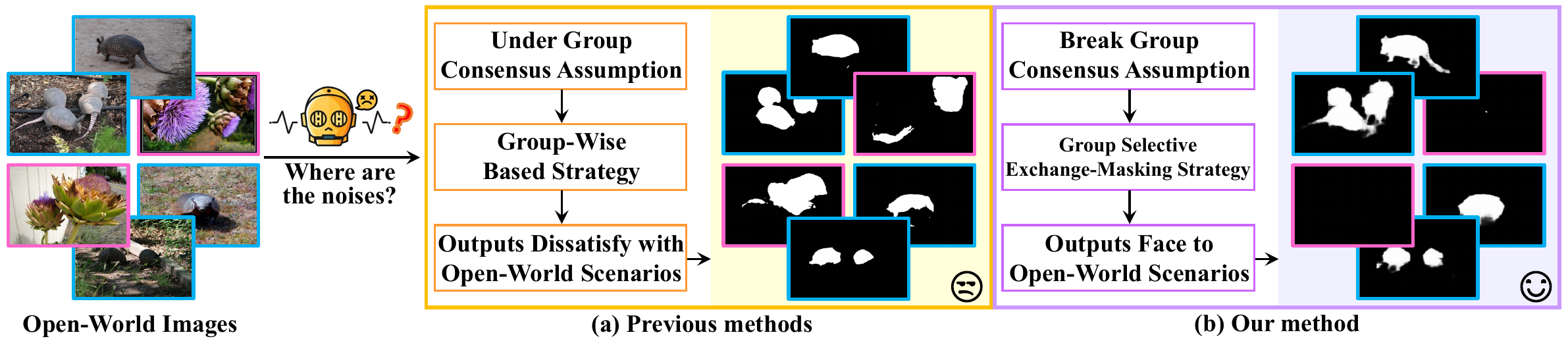}
\end{tabular}
\end{center}
\caption{
The group consensus assumption greatly affects the application of the model in the open world. When there exist irrelevant images in the test group, as shown in (a), previous works~\cite{zhang2020deep,fan2021group,zhang2020gradient,zhang2021summarize,zhang2022deep,yu2022democracy,wu2023group} customarily produce false positive predictions for the noisy image. However, helped by the designed group exchange-masking strategy, our model can achieve accurate predictions as shown in (b).
}
\label{fig: introduction}
\end{figure*}
\begin{figure*}[!t]
\begin{center}
\begin{tabular}{c}
\hspace{-0.18cm}\includegraphics[width=1\textwidth]{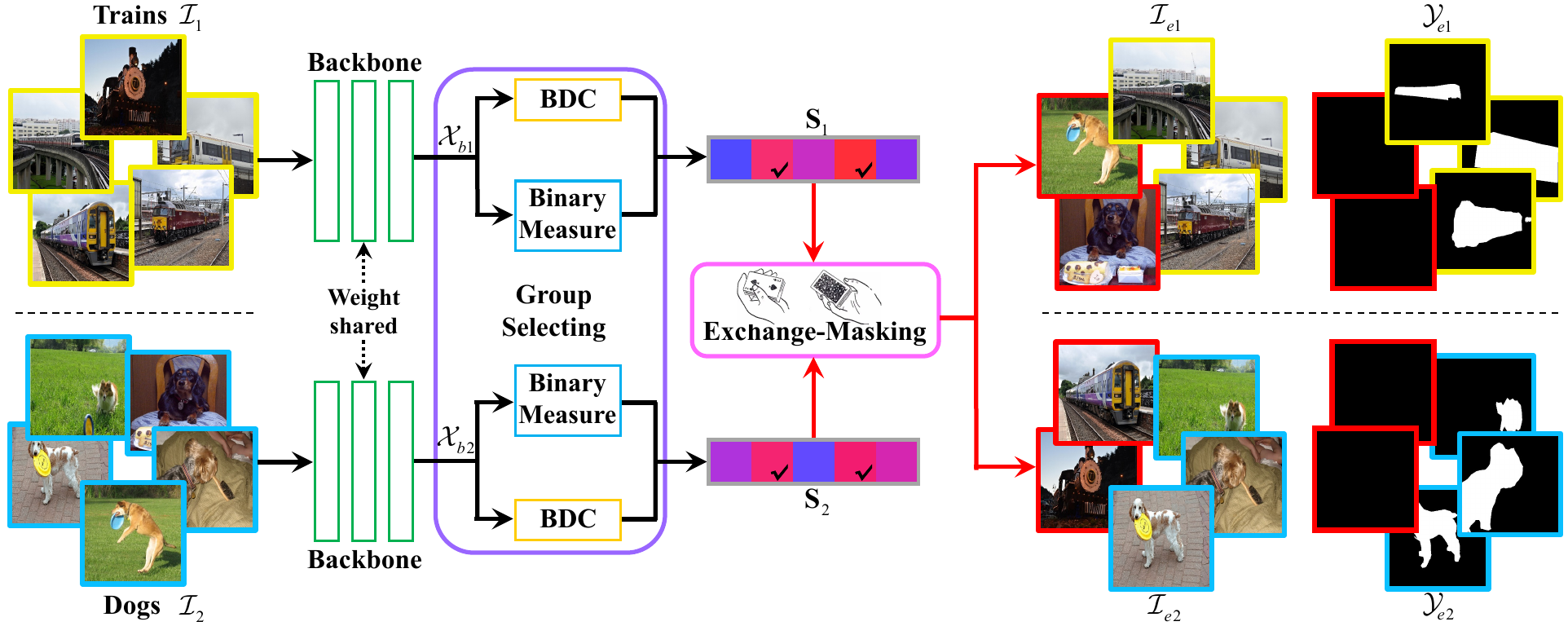}
\end{tabular}
\end{center}
\caption{
Given two image groups that contain different types of co-salient objects, we exchange several images between one group and the other. To enhance the effectiveness of robustness learning in our model, we select the most challenging images by using the BDC measure and the binarization measure. The selected `noisy' images are exchanged with the other group and the relevant ground truths are masked by all-zero maps.
}
\label{fig: GSEM}
\end{figure*}

%
In our work, we discover that the group consensus assumption can impose limitations on the robustness of CoSOD models when presented with images lacking a common object. As depicted in Figure~\ref{fig: introduction}(a), previous CoSOD models often produce false positive predictions for noisy images. This problem poses a barrier to the application of CoSOD models in open-world scenarios, where the actual inputs may include no-co-salient images.
To improve the model's robustness, we propose a learning framework with Brownian distance covariance (BDC)~\cite{xie2022joint} called group selective exchange-masking (GSEM). The GSEM is depicted in Figure~\ref{fig: GSEM}
%
%
When dealing with two sets of images that involve different semantic categories of co-salient objects, we perform an exchange of several images between the two groups, resulting in what we refer to as ``noisy images''. 
The quantity of noisy images chosen and swapped is deliberately maintained at a lower level than the number of the remaining pertinent images within the group. This guarantees that the salient object within the noisy images is treated as a non-co-salient object rather than the predominant co-salient object.
%
To enhance the effectiveness of robustness learning in our model, we need to select the most challenging images as ``noise images''. Given the specificity of the CoSOD task, we believe that measuring the difficulty of images should encompass two aspects: (1) the difficulty in capturing high-dimensional no-linear semantic information in images and (2) the difficulty in segmenting low-dimensional pixel-level information in images. To accomplish this, we assess the challenge of accurately segmenting images by combining the BDC measure and the binarization measure. By exchanging the model's learned perception of the most challenging images, we can better acquire robust features.
The term ``masking'' strategy pertains to the re-labeling of these noisy images. Since there are no co-salient objects in them, the original ground-truth objects are masked in the regenerated labels. The main learning goal is to make precise predictions for both the co-salient objects in the initial related images and the newly exchanged noisy images.

%
%
%
Incorporating noisy images into the training image group introduces an element of uncertainty into the learning process of the CoSOD model, as there exists a probability that each image may not contain the expected common object. In situations where we still need to extract common group information, this can greatly confuse the model. 
To address this challenge, we have devised a parallel structure feature extraction strategy that not only captures the group consensus property but also models the group uncertainty.
Specifically, we introduce a latent variable generator branch (LVGB) to capture the uncertainty-based global image features. 
%
We design the LVGB with Vector Quantised-Variational Autoencoder (VQ-VAE)~\cite{van2017neural} which is based on generating discrete latent variables, these discrete latent variables align more closely with the data distribution in open-world scenarios and possess a more powerful unsupervised representation capability compared to the continuous latent variables in VAE~\cite{kingma2013auto,van2017neural}, they exhibit better robustness when confronted with unknown classes. 
The features generated by VQ-VAE can effectively encompass both global characteristics and randomness, exhibiting strong suppression of features from noisy images and correction of the overconfidence exhibited by related images during the training phase.
VQ-VAE is a commonly employed technique to handle uncertainty in various computer vision applications including semantic scene completion ~\cite{lee2023diffusion}, high-quality image generation~\cite{razavi2019generating} and robust model learning~\cite{wu2023co}. 
%
Running concurrently with the LVGB, we route the image group into a CoSOD Transformer Branch (CoSOD-TB). 
Inside the CoSOD-TB, we partition every image group into localized patches. The incorporation of an attention mechanism within the transformer empowers this branch to capture local features by leveraging inter-patch correlations.
%
Consequently, this branch is capable of capturing the group consistency information. The outputs of both LVGB and CoSOD-TB are combined and passed into a transformer-based decoder for the prediction of co-salient objects. 
This paper makes several significant contributions as follows:
\begin{itemize}
\item We design a robust CoSOD model learning framework for open-world scenarios with a group selective exchange-masking strategy. 
This contrasts with previous CoSOD model learning frameworks that employ sets of related images as training data.
\item 
We have designed a parallel feature extraction mechanism composed of LVGB and CoSOD-TB. LVGB excels at modeling the uncertainty within a set of images and generates global stochastic features. CoSOD-TB, on the other hand, is employed to capture the consensus feature in the relevant images.
%
\item We analyze and restructure the three most commonly used benchmark datasets, CoSal2015~\cite{zhang2015co}, CoCA~\cite{zhang2020gradient}, and CoSOD3k~\cite{fan2021re}. We introduced `noisy images' into each group of relevant images in the datasets to simulate situations frequently encountered in open-world scenarios, where a set of images may include unrelated ones. We name the newly proposed datasets OWCoSal, OWCoSOD, and OWCoCA.
\item Extensive experiments on CoSal2015~\cite{zhang2015co}, CoCA~\cite{zhang2020gradient}, and CoSOD3k~\cite{fan2021re} have demonstrated the effectiveness of our approach. Additionally, to assess the model's performance in open-world scenarios, we conducted experiments and comparisons on three open-world datasets OWCoSal, OWCoSOD, and OWCoCA with existing state-of-the-art models, further affirming that our method outperforms others even in open-world scenarios.

\end{itemize}

The composition of the following sections in this paper is as follows: In Section~\ref{sec2}, we introduce the related works of our methods. In Section~\ref{sec3}, we describe our method's framework and its key designs. Then, we present sufficient experimental results in Section~\ref{sec4}. At last, we summarize our work and explore possible improvements and application directions in Section~\ref{sec5}.

This work is extended from our previous CVPR2023 work CoGEM~\cite{wu2023co} and ICASSP2023 work GWCoST~\cite{wu2023group}. We have made many extensions in task assumption expansion, method enhancement, experimental analysis, dataset investigation and restructuring. 
Firstly, we provide a more comprehensive introduction and supplement more recent related works in the related work. Moreover, we provide a detailed description and explanation of the unreasonable assumptions made in the previous CoSOD works, outlining potential challenges that CoSOD models may encounter in open-world scenarios.
Secondly, we improve the selection criteria for `noisy images' in GSEM by utilizing BDC, which takes into account the high-dimensional non-linear relationships between images, greatly assisting the model in identifying genuinely challenging samples, and the improved design is named GSEM. 
Thirdly, to better model uncertainty, we improve the CVAE in LVGB with VQ-VAE, enabling LVGB to generate high-quality uncertainty features more effectively.
Fourth, we analyze and restructure the three most commonly used benchmark datasets to simulate situations frequently encountered in open-world scenarios. We hope the restructured datasets can assist the field in evaluating the robustness of the CoSOD model in the open world.

\section{Related Work}\label{sec2}
\subsection{Co-salient Object Detection}
Earlier CoSOD approaches often employ non-learning approaches such as manually designed feature operators to capture the connections in a set of images.
In~\cite{chang2011co}, the authors conduct handcrafted features extracted from images, including Gabor~\cite{gabor1946theory} or SIFT features~\cite{lowe2004distinctive}, and perform co-saliency detection by leveraging the consistency relationships among low-level features. In~\cite{li2014efficient}, a manifold ranking scheme-based approach is introduced to predict saliency maps for capturing constraints within the images. There are also methods that utilize global contextual information obtained through clustering~\cite{fu2013cluster} or transnational alignment techniques~\cite{jacobs2010cosaliency} and applied it to co-saliency detection. After this, Jiang $et \ al.$~\cite{jiang2020co} use intermediate features to handle this task which include the results of saliency detection or image segmentation.

In more recent times, there has been a notable increase in learning-based CNN CoSOD models by learning network parameters to perform feature representation and output prediction.~\cite{wei2017group,hsu2018co,liu2023self,zhang2019co,zhang2020gradient,zhang2020adaptive,fan2021group}.
Wei $et \ al.$\cite{wei2017group} adopt the group consistency assumption as prior information and design the model's input and output in the form of groups to capture both the inherent features of images and the relationships between images.
Hsu $et \ al.$ and Liu $et \ al.$~\cite{hsu2018co,liu2023self} attempt to design an unsupervised or further self-supervised framework for co-salient object detection.
In~\cite{zhang2019co}, Zhang $et \ al.$ introduce a hierarchical manner for CoSOD in which the initial predicted results can undergo refinement through label smoothing to get more accurate results.
Zhang $et \ al.$~\cite{zhang2020gradient} explore the knowledge in the gradient during model training and utilize it to help the model focus on co-salient regions.
The method~\cite{zhang2020adaptive} designs a CNN-graph model for CoSOD to capture the inter- and intra-image information.
Fan $et \ al.$~\cite{fan2021group} propose a siamese network to capture and interact information between two groups. 
With the success of vision transformers(ViT)~\cite{dosovitskiy2020image}, a wave of research utilizing transformers has also started to emerge~\cite{wu2023group,wu2023co,wu2023object,su2023unified}. In~\cite{wu2023group}, a BDC module is proposed to help the CoSOD model capture no-linear information to be more discriminative. 
Wu $et \ al.$~\cite{wu2023object} introduce physical prior information from depth maps to assist the model in obtaining more accurate results.
Su $et \ al.$~\cite{su2023unified} design a unified framework for several works that all need group information by utilizing transformer technology.

\subsection{Robust Model and Feature Learning}
With the advancement and application of deep learning, researchers have observed a drop in the performance of models trained on closed datasets when applied to real-world open scenarios. Hence, there is increasing attention on how to enhance the robust model and feature learning, which is considered a potential pathway to overcoming the bottleneck~\cite{pinto2017robust}.
%
Numerous techniques have been introduced to enhance the model's robustness.
%
In \cite{christiano2016transfer} and \cite{rusu2017sim}], methods for transferring learned policies from a simulator to the real world are presented as a means to enhance feature robustness.
Heess $et \ al.$\cite{heess2015memory} adopt a reinforcement learning approach and design the recurrent neural networks to perform direct adaptive prediction. 
Rajeswaran $et \ al.$~\cite{rajeswaran2016epopt} acquire a resilient policy through the sampling of worst-case trajectories from a set of parameterized models to learn robust policies.
Liu $et \ al.$~\cite{liu2018towards} introduce an innovative defense algorithm named by merging randomness and ensemble which achieves a good improvement in robustness.

In addition to the design of the model, there have also been many related developments at the data level.
Xie $et \ al.$~\cite{xie2017mitigating} apply random elementary data augmentations to the images input to the model, including resizing and padding. They use these enhanced images to train the model and achieve improvements in model robustness.
After the work of Xie $et \ al.$~\cite{xie2017mitigating}, mixup\cite{zhang2017mixup} is proposed to improve the generalization and robustness of models, this has inspired many other approaches, including \cite{verma2019manifold}, \cite{yun2019cutmix}, \cite{kim2020puzzle} and \cite{kim2021co}. They all preprocess the input data to construct new training samples, and the samples they worry about have no targeted antagonistic information.
Defending against adversarial examples is also a new way to improve the robustness of models and attracted a lot of attention.
One common approach involves training the model using adversarial images. Wang $et \ al.$\cite{wang2019improving} incorporate mispredicted examples in adversarial training as a regularizer to improve the robustness of the model. 
Adversarial training is effective against attacks and further improves the robustness of the model. However, generating adversarial examples during the training process can be computationally expensive. Several studies have proposed techniques to mitigate the computational cost~\cite{shafahi2019adversarial,zhang2019you,wong2020learning}.

\subsection{Variational Deep Probabilistic Models}
Variational autoencoder (VAE)~\cite{kingma2013auto} and its conditional variational autoencoder (CVAE)~\cite{sohn2015learning} are used to generate latent variables from latent space.
In an earlier period, VAE and CVAE are widely used in zero-shot learning~\cite{mishra2018generative}, structured sequence prediction~\cite{bhattacharyya2019conditional}, image background modeling~\cite{li2019supervae}. 
More recently, VAE and CVAE have been adopted in various vision tasks, including RGB-D saliency detection\cite{zhang2021uncertainty}, action quality assessment\cite{zhou2022uncertainty}, image reconstruction\cite{zhang2021conditional}. 
Unlike VAE, which learns a continuous distribution, VQ-VAE~\cite{van2017neural} aims to learn a discrete representation by compressing images into a discrete latent space. Due to its high generation quality, VA-VAE has received considerable attention in computer vision, including video generating~\cite{yan2021videogpt}, image inpainting~\cite{peng2021generating}, robust semantic learning~\cite{hu2023robust}, to name a few.
\begin{figure*}[!t]
\hspace{-0.18cm}\includegraphics[width=1\textwidth]{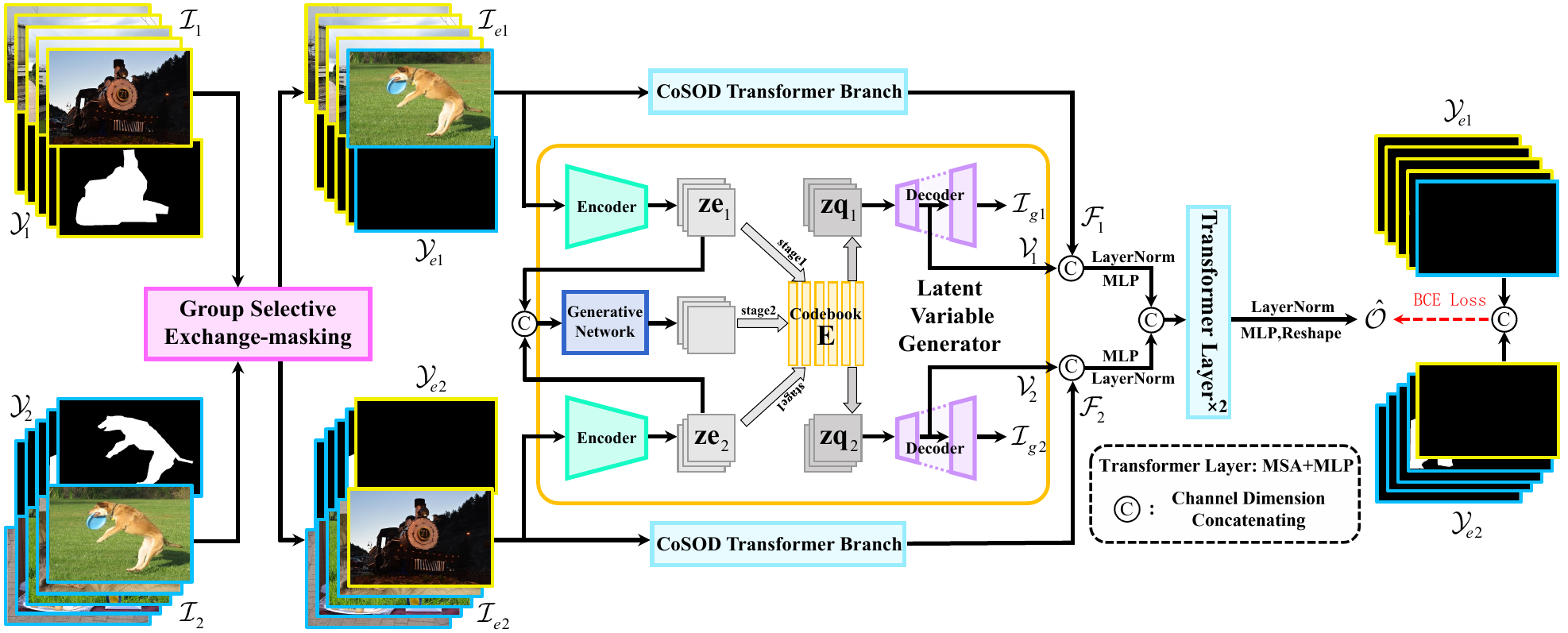}
\caption{
The overall flowchart of our proposed CoGSEM. Two image groups $\{\mathcal{I}_{1},\mathcal{I}_{2}\}$ with their masks $\{\mathcal{Y}_{1},\mathcal{Y}_{2}\}$ are first rearranged and generated by the group selective exchange-masking strategy, yielding the exchanged image groups $\{\mathcal{I}_{e1},\mathcal{I}_{e2}\}$ and their corresponding masked labels $\{\mathcal{Y}_{e1},\mathcal{Y}_{e2}\}$. 
Then, $\mathcal{I}_{e1}$ and $\mathcal{I}_{e2}$ are fed to the latent variable generator, generating the uncertainty-based features $\{\mathcal{V}_{1},\mathcal{V}_{2}\}$ with global clues to eliminate the bias caused by noisy images.
At the same time, $\{\mathcal{I}_{e1},\mathcal{I}_{e2}\}$ are fed to two  CoSOD transformer branches with the parallel architecture and shared weights, outputting feature sequences $\{\mathcal{F}_{1},\mathcal{F}_{2}\}$ with group consistency information and long-range independent information. 
%
In the end, $\mathcal{F}$ and $\mathcal{V}$ are concatenated along the channel dimension and passed through the decoder based on a transformer structure to obtain the final predicted co-saliency maps $\hat{\mathcal{O}}$.
}
\label{fig: flowchart}
\end{figure*}

\section{Proposed method}\label{sec3}
Our framework's pipeline is depicted in Figure~\ref{fig: flowchart}. 
Two groups of $2N$ images $\mathcal{I}_{i}=\{I_i^n\in \mathbb{R}^{H\times W\times 3}\}_{n=1}^N,i=1,2$ and the corresponding manually-labeled binary masks $\mathcal{Y}_{i}=\{\textbf{Y}_i^n\in \mathbb{R}^{H\times W}\}_{n=1}^N$ with different categories are given as input, the GSEM strategy is utilized to process both $\mathcal{I}_{i}$ and $\mathcal{Y}_{i}$, yielding the exchanged $\mathcal{I}_{ei}$ and generated $\mathcal{Y}_{ei}$.
Each group of images $\mathcal{I}_{ei}$ is fed to the LVGB which is based on VQ-VAE. 
In LVGB, we first obtain compressed image features $\textbf{ze}_{i}\in \mathbb{R}^{N\times C\times H\times W}$ through a VQ-VAE encoder. These features are then passed through our designed generative network to generate results, which are sampled from the codebook $e\in\mathbb{R}^{K\times D}$ using a nearest-neighbor approach to obtain $\textbf{ze}_{i}\in \mathbb{R}^{N\times C\times H\times W}$. 
Subsequently, we use the VQ-VAE decoder to reconstruct the generated image $\mathcal{I}_{gi}=\{I_{gi}^n\in \mathbb{R}^{H\times W\times 3}\}_{n=1}^N,i=1,2$. During the intermediate stages of the VQ-VAE decoder, 
we further extract a latent feature sequence $\mathcal{V}_i$, which will be resized in subsequent steps to have the same dimensions as the features obtained from the CoSOD-TB.
%
At the same time, the image groups are fed into the CoSOD-TB, which has a Siamese structure with shared weights, and we obtain the feature sequence $\mathcal{F}_i$. $\mathcal{F}_i$ captures group consensus and the long-range dependency information and has strong representational power.
Subsequently, $\mathcal{V}_i$ and $\mathcal{F}_i$ are concatenated along the channel dimension and then processed by several subsequent processes, as indicated in Figure~\ref{fig: flowchart}. This is followed by an up-sampling layer, resulting in the corresponding predicted co-saliency maps $\hat{\mathcal{O}}=\{\hat{\textbf{O}}^n\in \mathbb{R}^{H\times W}\}_{n=1}^{2N}$.
%
%
\begin{algorithm}[!t]
\caption{Group selective exchange-masking}
\hspace*{0.02in} {\bf Input:}
$\mathcal{I}_{1}=\{I_1^n\}^N_{n=1}, \mathcal{Y}_{1}=\{\textbf{Y}_1^n\}^N_{n=1}$,\\ \hspace*{0.50in} $\mathcal{I}_2=\{I_2^n\}^N_{n=1}, \mathcal{Y}_2=\{\textbf{Y}_2^n\}^N_{n=1}.$\\
\hspace*{0.02in} {\bf Output:}
$\mathcal{I}_{e1}$, $\mathcal{Y}_{e1}$, $\mathcal{I}_{e2}$, $\mathcal{Y}_{e2}.$

\begin{algorithmic}[1]
    \STATE $\mathcal{I}_{e1}$, $\mathcal{I}_{e2}\leftarrow$ Rearranging $\mathcal{I}_1$, $\mathcal{I}_2$ via group selective stage.
    \STATE $\mathcal{Y}_{e1}$, $\mathcal{Y}_{e2}\leftarrow$ Rearranging $\mathcal{Y}_1$, $\mathcal{Y}_2$ via exchange-masking stage.
    \FOR{$n=1,2,\ldots, k$}
    \STATE $I_{e1}^n\leftarrow I_2^n,I_{e2}^n\leftarrow I_1^n$,
    \STATE $\textbf{Y}_{e1}^n\leftarrow \textbf{0}\in \mathbb{R}^{H\times W\times 1}$,
    \STATE $\textbf{Y}_{e2}^n\leftarrow \textbf{0}\in \mathbb{R}^{H\times W\times 1}$.
    \ENDFOR
\end{algorithmic}
\label{alg: exchange}
\end{algorithm}

\subsection{Group Selective Exchange-Masking}

The widely adopted group consensus assumption in CoSOD models and datasets significantly constrains the model's robustness. This limitation becomes particularly apparent in open-world scenarios where the test image group includes images that lack co-salient objects. To address this, we restructure the two input image groups using the GSEM strategy tailored for CoSOD.
Algorithm~\ref{alg: exchange} summarises the procedure of GSEM: taking raw $\mathcal{I}_i$ and $\mathcal{Y}_i$ as input, the goal of GSEM is to produce $\mathcal{I}_{ei}$, which includes noisy images, and the corresponding all-zero masked labels in $\mathcal{Y}_{ei}$. 
%
%
Specifically, as shown in Figure~\ref{fig: GSEM}, we first crop $\mathcal{I}_i$ and $\mathcal{Y}_i$ into patches as inputs, after the processing in the transformer 
backbone~\cite{yuan2021tokens} we can get the token sequence $\mathcal{X}_{b1},\mathcal{X}_{b2}\in\mathbb{R}^{N\times \frac{H}{16}\times \frac{W}{16}\times c}$. Next, we design a novel metric to measure the difficulty of images. Considering the specificity of the CoSOD, the metric is constructed in a mixed manner to consider both the difficulty in capturing high-dimensional no-linear semantic information in images and the difficulty in segmenting low-dimensional pixel-level information in images. We adopt the BDC and the binary measure together in the group selective stage. 

In the designed BDC, the token sequences $\mathcal{X}_{b1}, \mathcal{X}_{b2}\in\mathbb{R}^{N\times \frac{H}{16}\times \frac{W}{16}\times c}$ are compressed to $\mathcal{X}_{g1}, \mathcal{X}_{g2}\in\mathbb{R}^{1\times \frac{H}{16}\times \frac{W}{16}\times c}$ to represent the group consistency information.
%
%
Take one branch as example, $\mathcal{X}_{g1}$ and $\mathcal{X}_{b1}^n, n=1,2,...,N$ represent the values of the random variables $M$ and $U$. 
Consequently, we can use the $\phi_{MU}$ to describe the nonlinear interactions between $M$ and $N$ when considering the joint distribution state.
We define the BDC metric $\rho$ in the same form as~\cite{xie2022joint,wu2023group}
\begin{equation}
\rho(M,U) = \int_{\mathbb{R}_p}\int_{\mathbb{R}_q}\frac{|\phi_{MU}(\textbf{t},\textbf{s})-\phi_{M}(\textbf{t})\phi_{U}(\textbf{s})|^2}{d_p d_q ||\textbf{t}||^{1+p}||\textbf{s}||^{1+q}}d\textbf{t}d\textbf{s}, 
\label{eq: BDC}
\end{equation}
where $\phi(\cdot)$ is the joint characteristic function, $||\cdot||$ denotes Euclidean norm, $p = q = \frac{H}{16}\times \frac{W}{16}\times c$, $d_p=\pi^{(1+p)/2}/\Gamma((1+p)/2)$, and $\Gamma$ is the complete gamma function.

For a group of $c$ observations $\{(\textbf{m}_1,\textbf{u}_1)...(\textbf{m}_c,\textbf{u}_c)\}$ sampled in the channel dimension from $\mathcal{X}{g1}$ and $\mathcal{X}{b1}^{n}$ with the characteristics of independent and identical distribution(i.i.d), $\phi{MU}$ can be defined as
\begin{equation}
\phi_{MU}(\textbf{t},\textbf{s})=\frac{1}{c}\sum_{k=1}^{c}
\exp(i(\textbf{t}^{\top}\textbf{m}_{k}+\textbf{s}^{\top}\textbf{u}_{k})),
\end{equation}
where $i$ is the imaginary unit and $\top$ denotes matrix transpose. 

For discrete data, we can rewrite~(\ref{eq: BDC}) using an approximate form with the following processes.
We define $\hat{\textbf{A}}=(\hat{a}_{kl})\in \mathbb{R}^{c\times c}$, where $\hat{a}_{kl}=||\textbf{m}_{k}-\textbf{m}_{l}||$ is the Euclidean distance between a pair of observations in $\mathcal{X}_{g1}$. $\hat{\textbf{B}}=(\hat{b}_{kl})\in \mathbb{R}^{c\times c}$, $\hat{b}_{kl}=||\textbf{u}_{k}-\textbf{u}_{l}||$. Then, (\ref{eq: BDC}) can be re-written in a simple form as\cite{szekely2009brownian,xie2022joint}
\begin{equation}
\rho(\mathcal{X}_{g1},\mathcal{X}_{b1}^{n})=tr(\textbf{A}^{\top}\textbf{B}),
\end{equation}
where $tr(\cdot)$ means matrix trace, $\textbf{A}=(a_{kl})$ is BDC matrix and $a_{kl}=\hat{a}_{kl}-\frac{1}{c}\sum_{k=1}^c\hat{a}_{kl}-
\frac{1}{c}\sum_{l=1}^c\hat{a}_{kl}-\frac{1}{c^2}\sum_{k=1}^c\sum_{l=1}^c\hat{a}_{kl}$, the matrix $\textbf{B}$ is computed in the same way . Since the BDC matrix is symmetric, $\rho(\mathcal{X}_{g1},\mathcal{X}_{b1}^{n})$ can be further expressed in a simple form as the inner product of two BDC vectors $\textbf{a}$ and $\textbf{b}$ as
\begin{equation}
\rho(\mathcal{X}_{g1},\mathcal{X}_{b1}^{n})=\textbf{a}^{\top}\textbf{b},
\end{equation}
%
we derive $\textbf{a}$ by extracting the upper triangular elements of $\textbf{A}$ and subsequently vectorizing them, $\textbf{b}$ is obtained by the same way.

The high-dimensional no-linear group difficulty measured by BDC $\textbf{s}^{BDC}\in \mathbb{R}^{N\times 1}$ in group one is formulated as 
\begin{equation}
\textbf{s}^{BDC}_{1} = \{\rho(\mathcal{X}_{g1},\mathcal{X}_{b1}^{n})\}_{n=1}^{N}.
\end{equation}

In the designed binary measure, we first resize the token sequence $\mathcal{X}_{b1}\in\mathbb{R}^{N\times \frac{H}{16}\times \frac{W}{16}\times C}$ to $\mathcal{X}_{bin1}\in\mathbb{R}^{N\times \frac{H}{16}\times \frac{W}{16}\times 1}$. The feature from the backbone contains rich positioning information and preliminary target shapes~\cite{zhou2016learning}, we believe that the information contained in these early, unprocessed features is sufficient to measure the model's grasp of the samples. 
To this end, we perform element-wise Hadamard product between the feature map $\mathcal{X}_{bin1}^{n}$ and the resized ground truth $\mathcal{Y}_{bin1}^{n}\in\mathbb{R}^{1\times \frac{H}{16}\times \frac{W}{16}\times 1}, n=1,..., N$ to obtain a measure of consistency.
\begin{equation}
\textbf{s}^{Bin}_{1} = \{\sum_{j}(\mathcal{X}_{bin1}^{n} \odot \mathcal{Y}_{bin1}^{n})\}_{n=1}^{N},
\end{equation}
where $j\in\mathbb{R}^{\frac{H}{16}\times \frac{W}{16}}$ is the pixel from the feature map, and $\textbf{s}^{Bin}_{2}$ can be obtained by the same manner.

The difficulty index $\textbf{s}$ is defined as follows
\begin{equation}
\textbf{s}_{1} = \textbf{s}^{BDC}_{1} + \mu \textbf{s}^{Bin}_{1},
\label{eq: bdc}
\end{equation}
where $\mu$ is the hyper-parameters to balance the weight.

%
We select the top-$k$ images from each group based on $\textbf{s}_{i}, i=1,2$, where these $k$ images represent the most challenging samples within each group. These selected $k$ images are then exchanged between the two groups, ensuring that the chosen images possess a high level of difficulty and enhance the effectiveness of the training process.
The training process is fundamentally a mini-max optimization like an adversarial training process~\cite{madry2018towards} as
\begin{equation}
\label{eq:minmaxloss}
\mathop{\min} \limits_{\boldsymbol{\varphi}}\sum^k_{n=1}\sum^2_{i=1}\mathop{\max} \limits_{\mathcal{I}_i(n,:)}\mathcal{L}(f_{\boldsymbol{\varphi}}(\mathcal{I}_i(n,:)),\hat {\mathcal{O}}_i(n,:)),
\end{equation}
where the loss $\mathcal{L}$ is defined by (\ref{eq:loss_total}), and $f$ denotes the entire model with learnable parameters $\boldsymbol{\varphi}$.
%
Solving (\ref{eq:minmaxloss}) enables the identification of images with the highest noise levels, thereby maximizing the training loss. Our aim is to minimize the loss function with respect to the intra-group noisy images to enhance the robustness of the model.
%
At last, we replace the labels of images containing significant noise with maps of all zeros.
%
%
%
%
\subsection{Latent Variable Generator Branch}
\begin{figure*}[!t]
\begin{center}
\begin{tabular}{c}
\hspace{-0.20cm}\includegraphics[width=1\textwidth]{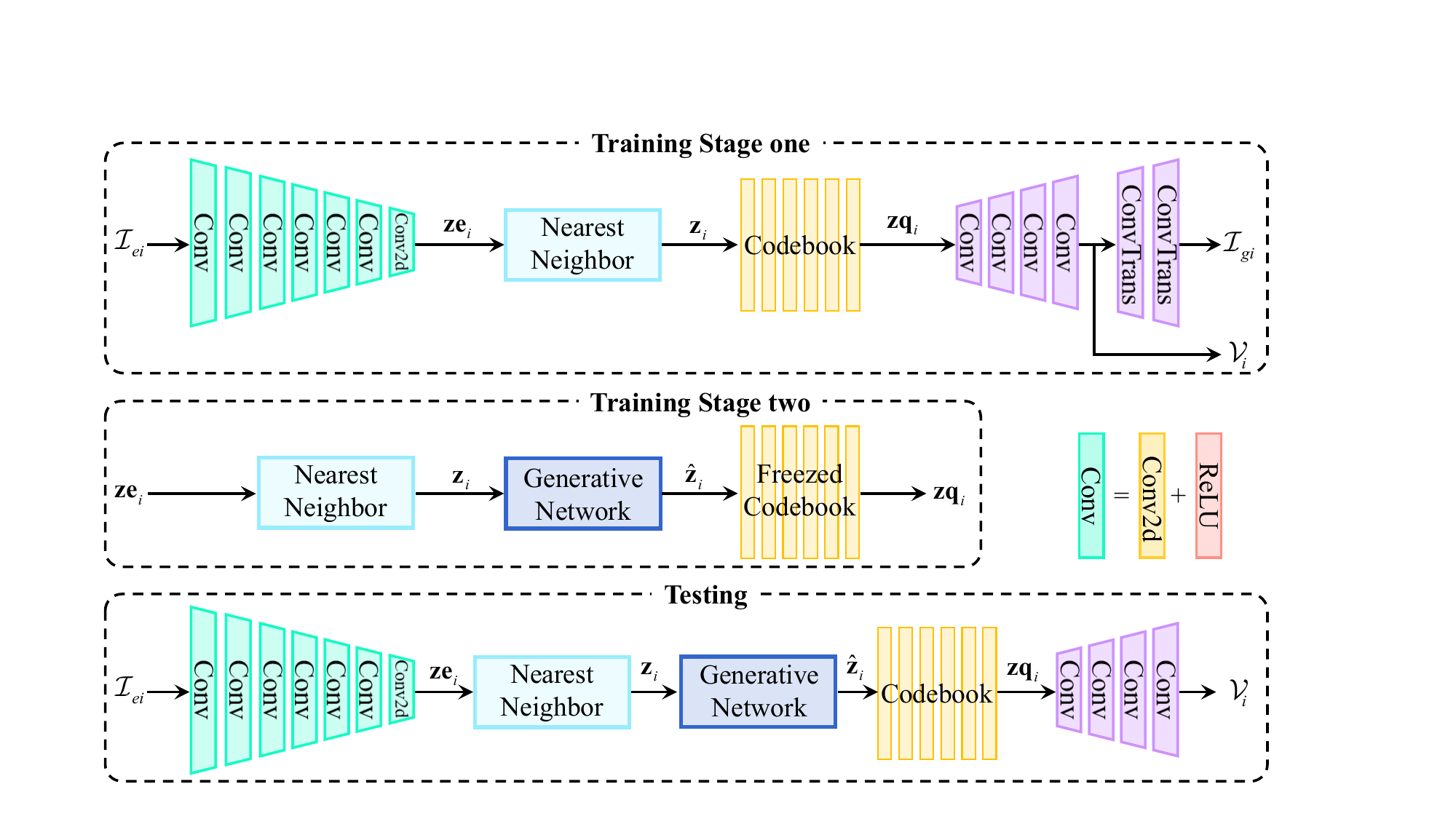}
\end{tabular}
\end{center}
\caption{Architecture of the LVGB.}
\label{fig: LVGB}
\end{figure*}
With the assistance of GSEM, by training the model with image groups containing noisy images, our model's robustness can be significantly improved, enabling it to identify whether a single image contains co-salient objects.
%
Nevertheless, due to the uncertainty introduced by incorporating noisy images into the training image groups, the model is disturbed and tends to focus on the non-co-salient regions. Consequently, during the initial stages of training, the model becomes overconfident in background regions or erroneously identified objects, resulting in inaccurate predictions.

To tackle this problem, we introduce LVGB as a solution. LVGB learns to encode down-scaled latent codes of high-resolution inputs while maintaining a discrete codebook. This approach allows us to denoise the redundancies and capture the most vital consensus information within an image group~\cite{kohl2018probabilistic,yan2021videogpt}.
%
The discrete variables sampled from the codebook effectively represent the uncertainty of group consensus information while preserving group-common characteristics. These variables can be utilized to modulate the deterministic features from other branches, thereby highlighting co-salient objects and suppressing noise samples.

Specifically, we train a VQ-VAE~\cite{van2017neural} and a generative network in two stages. To learn discrete latent encodes $\textbf{zq}_{i}$, we first train a VQ-VAE on group data. VQ-VAE consists of three main components: an encoder $f_{VQ-encoder}$, a codebook $\textbf{E}=\{\textbf{e}_{k}\in\mathbb{R}^{1\times D}\}_{k=1}^K$, and a decoder $f_{VQ-decoder}$. As shown in Figure~\ref{fig: LVGB}, the encoder consists of some convolution and activation operations, which process the input $\mathcal{I}_{ei}$ as $\textbf{ze}_{i}$,
\begin{equation}
\textbf{ze}_{i} = f_{VQ-encoder}(\mathcal{I}_{ei}).
\end{equation}
Next, we calculate the distance between $\textbf{ze}_{i}$ and the embeddings in the codebook $\textbf{E}$ using the nearest-neighbor algorithm to obtain the corresponding index $\textbf{z}_{i}$. Subsequently, we retrieve the discrete latent variable $\textbf{zq}_{i}$ from the codebook corresponding to $\textbf{z}_{i}$,
\begin{equation}
\textbf{zq}_{i} = Quantize(\textbf{ze}_{i}) = \textbf{e}_k, k = \mathop{\arg\min} \limits_{j}||\textbf{ze}_{i} - \textbf{e}_j||_2.
\end{equation}
When getting $\textbf{zq}_{i}$, we can obtain reconstructed $\mathcal{I}_{gi}$ through decoder
\begin{equation}
\mathcal{I}_{gi} = f_{VQ-decoder}(\textbf{zq}_{i}).
\end{equation}
The loss function used to train the VQ-VAE is formulated as
%
\begin{equation}
\begin{aligned}
\mathcal{L}_{VQ-VAE}&=\frac{1}{2N}\sum_{n=1}^{N}\{\sum_{i=1}^2\ell_{MSE}(\mathcal{I}_{ei}(n:,),\mathcal{I}_{gi}(n:,))\\
&+\ell_{MSE}(sg[\textbf{ze}_{i}],\textbf{zq}_{i})+\lambda_0 \ell_{MSE}(sg[\textbf{zq}_{i}],\textbf{ze}_{i})\},
\label{eq: vq}
\end{aligned}
\end{equation}
where $\lambda$ is the hyperparameter to balance the loss, $sg$ means stop gradient, and $\mathcal{L}_{VQ-VAE}$ is composed of three parts. The first component is the reconstruction loss, which encourages VQ-VAE to learn the ability to accurately recover features. The second component is the codebook loss, which minimizes the distance between the embedded $\textbf{zq}_{i}$ in the codebook and the input $\textbf{ze}_{i}$. The third component is the commitment loss, which prevents excessive fluctuations between different codes. 
The Mean-Squared Error (MSE) loss is defined as
\begin{equation}
\ell_{MSE}(\mathcal{I}_{ei}(n,:),\mathcal{I}_{gi}(n,:))=||\mathcal{I}_{ei}(n,:)-\mathcal{I}_{gi}(n,:)||^2_2,
\end{equation}

After training VQ-VAE, we acquire the capability to compress and reconstruct images. However, this is merely a reconstruction process and does not possess the ability to generate uncertainty representations. To achieve this, we need to train a generative network. Inspired by PixelCNN~\cite{razavi2019generating} and following its architecture, we designed a generative network, that consists of several residual gated convolution layers and casual multi-mead attention layers, it aims to sample new features from the probability distribution of input features, introducing uncertainty while preserving the primary features, thereby enhancing the model's robustness by mitigating overconfidence. The loss function used to train the generative network is as follows
\begin{equation}
\mathcal{L}_{GENER}=\ell_{CE}(\textbf{z}_{i},\hat{\textbf{z}}_{i}),
\label{eq: GENER}
\end{equation}
where $\ell_{CE}$ is a Cross-Entropy (CE)~\cite{rubinstein2004cross} loss defined as
\begin{equation}
\ell_{CE}(\textbf{z}_{i},\hat{\textbf{z}}_{i})=-\frac{1}{K}\sum_{k=1}^{K}\textbf{z}_{i}(k,:)log(\hat{\textbf{z}}_{i}(k,:)).
\label{eq: GENER}
\end{equation}

When testing, the data is followed as shown in Figure~\ref{fig: LVGB}.

\subsection{CoSOD Transformer Branch}

\begin{figure}[!t]
\begin{center}
\begin{tabular}{c}
\hspace{-0.20cm}\includegraphics[width=1.0\textwidth]{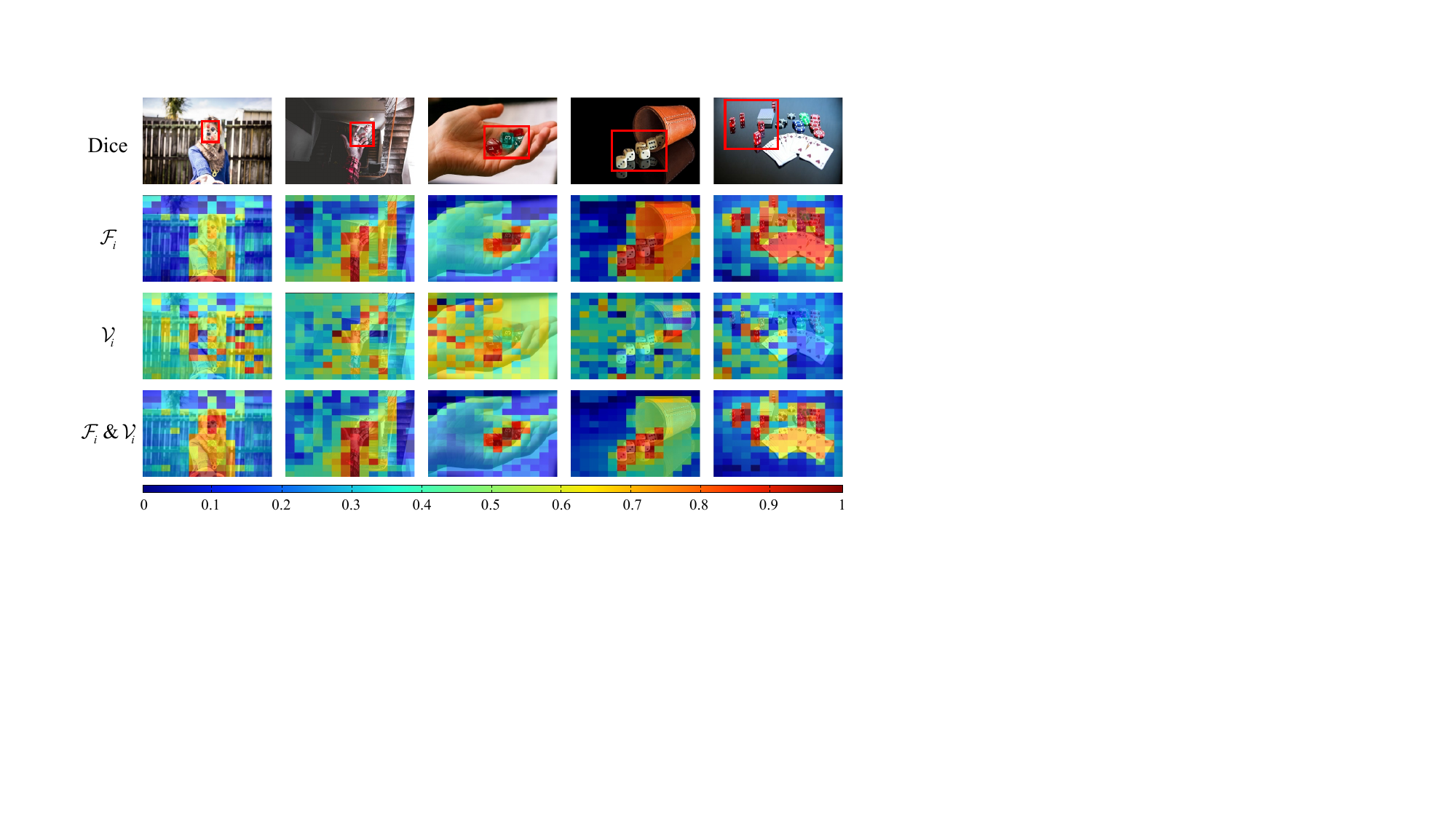}
\end{tabular}
\end{center}
\caption{Visualization of the effectiveness of stochastic features. The top row consists of images selected from the "Dice" group. The second row shows visualizations of the features $\mathcal{F}_{i}$ obtained from the CoSOD-TB. The third row displays the visualizations of the stochastic features $\mathcal{V}_{i}$ generated by the LVGB. The fourth row illustrates the visualizations of the features after integration during the transformer decoding phase.}
\label{fig: uncertainly}
\end{figure}

%
Once we obtain the stochastic features $\mathcal{V}_i$ from LVGB, we utilize these features to modulate the generation of the general features $\mathcal{F}_i$ within the CoSOD-TB. This process helps concentrate on the co-salient regions.
\begin{figure*}[!t]
\includegraphics[width=1.0\textwidth]{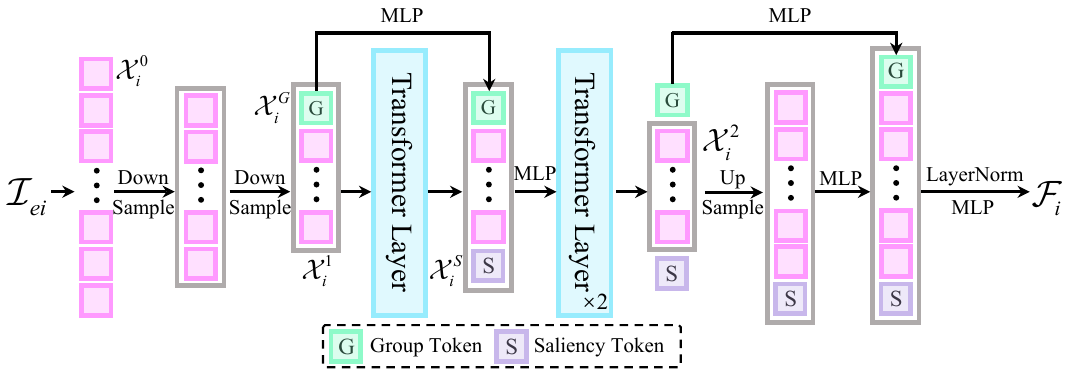}
\caption{Architecture of the CoSOD-TB.}
\label{fig: transformer}
\end{figure*}
In particular, as depicted in Figure~\ref{fig: transformer}, each group of input images, denoted as $I^n_i\in\mathcal{I}_{ei}$, is cropped into a total of $d$ image patches with dimensions $H/4\times W/4$. These image patches are stacked along the channel dimension to create the feature sequence $\mathcal{X}_{i}^{0}\in\mathbb{R}^{N\times \frac{H}{4}\times \frac{W}{4}\times 3d}$. Subsequently, the feature sequence is fed into a transformer-based feature extraction backbone with a T2T architecture~\cite{yuan2021tokens}. The T2T architecture consists of folding and unfolding components~\cite{yuan2021tokens,liu2021visual}, which effectively interact with the local information within the input data $\mathcal{X}_{i}$. As a result, the output $\mathcal{X}_{i}^{1}\in\mathbb{R}^{N\times \frac{H}{16}\times \frac{W}{16}\times c}$ from the transformer backbone encodes both global and local features simultaneously.
Then, to effectively capture group-common features essential for localizing co-salient objects and object-specific features crucial for preserving object details, we introduced $\mathcal{X}^{G}\in\mathbb{R}^{N\times 1\times c}$ and $\mathcal{X}^{S}\in\mathbb{R}^{N\times 1\times c}$. These feature tokens employ self-attention mechanisms to facilitate information propagation~\cite{wu2023co}.
In order to facilitate a comprehensive interaction between group-common features and object-specific features, we concatenate $\mathcal{X}^{G}$ and $\mathcal{X}^{S}$ with $\mathcal{X}_{i}^{1}$ along the channel dimension at different stages and feed them into a Transformer Layer to obtain a new feature sequence. Specifically, due to the presence of substantial group common noise, $\mathcal{X}^{G}$ undergoes an additional MLP operation. The design of the transformer layer is inspired by the successful approaches in the Vision Transformer (ViT)~\cite{dosovitskiy2020image}. After several MLP and Transformer Layer operations, we split the $\mathcal{X}^{G}$ and $\mathcal{X}^{S}$ to up-sample the feature sequence and obtain $\mathcal{X}_{i}^{2}$. $\mathcal{X}^{2}_{i}$ is further up-sampled and processed by several processes shown in Figure~\ref{fig: transformer}, we can get the general features $\mathcal{F}_{i}$. 
Next, $\mathcal{F}_{i}$ and the stochastic variable $\mathcal{V}_{i}$ are concatenated along the channel dimension and undergo some operations as illustrated in Figure~\ref{fig: flowchart}. The resulting feature sequence is then fed into a decoder that consists of Transformer Layers and an MLP structure.
During the decoding phase, with the modulation of the random variable V, our model has a better chance of overcoming overconfidence in the early stages of training, considering a wider range of potential correct co-objects. 
Figure~\ref{fig: uncertainly} illustrates the visual comparison of feature visualization. From the figure, it can be observed that without the incorporation of randomly stochastic feature $\mathcal{V}_{i}$, $\mathcal{F}_{i}$ might erroneously emphasize non-co-salient objects. Without proper guidance, the model may fall into the trap of excessive over-confidence. When stochastic features are integrated, the model effectively corrects the overconfidence in incorrect targets and suppresses the focus on non-co-salient regions.
%
Finally, the output sequences undergo reshaping to generate the predicted co-saliency maps denoted as $\hat{\mathcal{O}}$.

The CoSOD-TB's loss function is defined as
\begin{equation}
\mathcal{L}_{TRANS}=\frac{1}{2N}\sum_{n=1}^{N}\sum_{i=1}^2\ell_{BCE}(\mathcal{Y}_{ei}(n,:),\hat{\mathcal{O}_i}(n,:)),
\label{eq: trans}
\end{equation}
where $\ell_{BCE}$ is a Binary Cross-Entropy (BCE) loss~\cite{paszke2019pytorch} defined as
\begin{equation}
\begin{aligned}
\ell_{BCE}(\mathcal{Y}(n,:),&\hat {\mathcal{O}}(n,:))=-(\mathcal{Y}(n,:)^\top\log(\hat {\mathcal{O}}(n,:))\\&-(1-\mathcal{Y}(n,:))^\top\log(1-\hat {\mathcal{O}}(n,:))).
\label{eq: bce}
\end{aligned}
\end{equation}
\subsection{Loss Function}
$\mathcal{L}_{VQ-VAE}$ is used to train the VQ-VAE at the first training stage and $\mathcal{L}_{GENER}$ is used at the second training stage. During the second training stage, the weights of the VQ-VAE are frozen and only the parameters of generative network are updated. Once the training for VQ-VAE and generative network are completed, they are integrated into the model. The whole model is supervised by $\mathcal{L}_{TRANS}$. The multi-task loss to be optimized is defined as:
\begin{equation}
\mathcal{L} = \lambda_1 \mathcal{L}_{VQ-VAE} + \lambda_2 \mathcal{L}_{GENER} + \lambda_3 \mathcal{L}_{TRANS},
\label{eq:loss_total}
\end{equation}
where $\lambda_1$, $\lambda_2$, $\lambda_3$ are the hyper-parameters to select the loss function at different stage.


\subsection{Reorganized Dataset}
\begin{figure*}[!t]
\begin{center}
\includegraphics[width=1.0\textwidth]{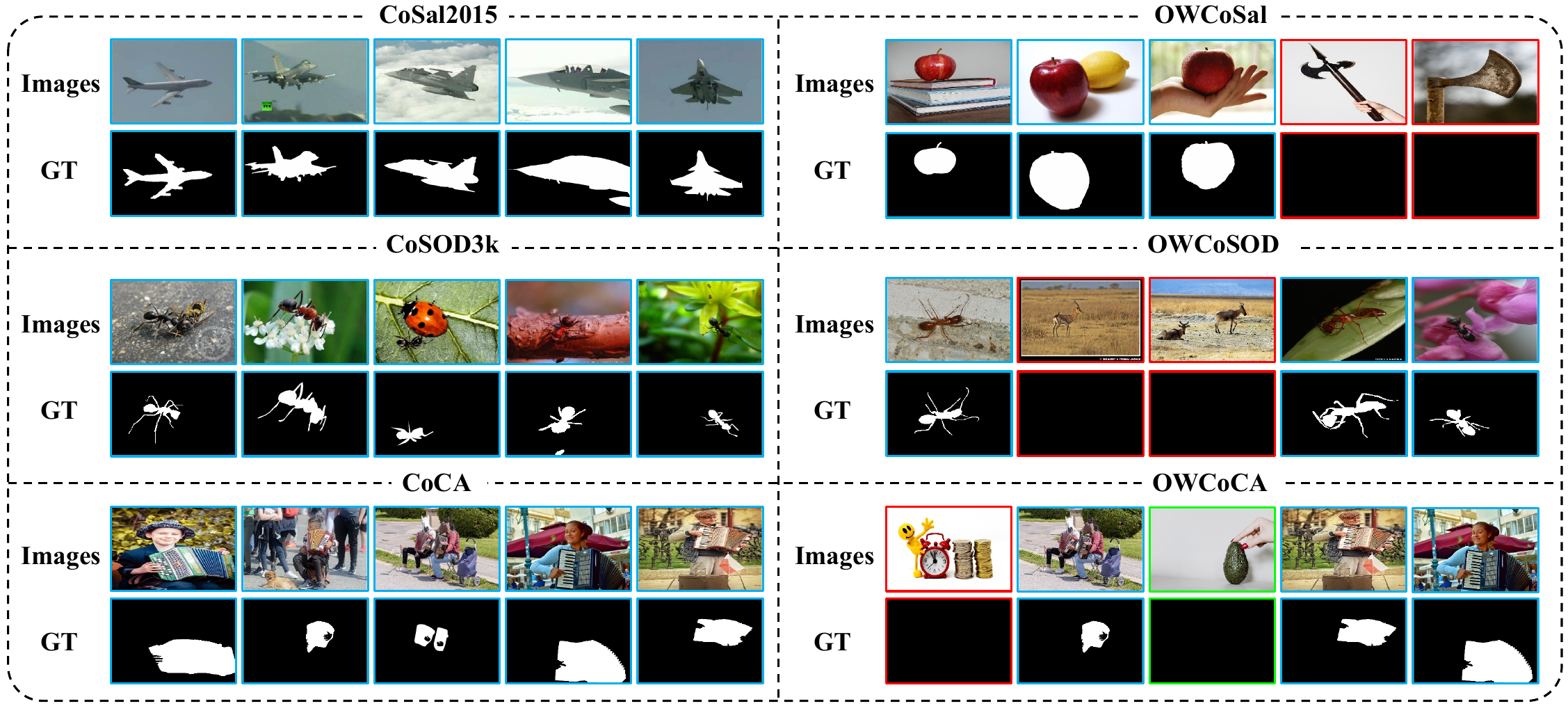}
\end{center}
\caption{Samples from six datasets. CoSal2015~\cite{zhang2015co}, CoSOD3k~\cite{fan2021re} and CoCA~\cite{zhang2020gradient} all follow the group consistency assumption, while our designed OWCoSal, OWCoSOD and OWCoCA contain different types of noise. }
\vspace{-0.4cm}
\label{fig: dataset}
\end{figure*}
In order to better simulate open-world scenarios, we conduct analysis and reorganization of the commonly used three largest and most challenging datasets: CoSal2015~\cite{zhang2015co}, CoSOD3k~\cite{fan2021re}, CoCA~\cite{zhang2020gradient}, and reorganize them to OWCoSal, OWCoSOD, OWCoCA.

We show the characters of existing datasets in the left half of Figure~\ref{fig: dataset}. One of the most commonly employed benchmark datasets is CoSal2015~\cite{zhang2015co}, consisting of 2,015 images distributed across 50 categories.
Its difficulty mainly stems from images within certain categories that have different backgrounds.
%
%
%
CoSOD3k~\cite{fan2021re} is currently the most extensive evaluation benchmark, featuring a total of 160 categories and more than 3000 images. In contrast to CoSal2015, CoSOD3k includes a significant number of images with two or three instances to be segmented, covering a wide range of Scenes, size variations, optical conditions, and backgrounds.
CoCA~\cite{zhang2020gradient} represents the most demanding evaluation benchmark, comprising 80 categories with a total of 1,295 images. This dataset includes many highly challenging samples, featuring objects from unknown categories, ghosted objects, extremely noisy backgrounds, and other challenging aspects.

Previously, researchers have constructed datasets based on the assumption of group consistency, leading to benchmark test sets structured by categories, with each category group containing images featuring the same salient object, \textit{i.e.}, co-salient objects. Researchers primarily designed the difficulty of test sets from the perspective of challenges in individual images, such as multiple scales, complex backgrounds, and interfering objects. To some extent, they have considered scenarios resembling open-world situations. However, they overlooked the consideration of the group consistency assumption itself. We believe that it is necessary to break this assumption. In open-world scenarios, the main challenge arises from unrelated images that do not contain co-salient targets. In application, collected sets of images will inevitably include noisy images, and the objective of CoSOD is to detect co-salient objects, even in the presence of a significant number of noisy images.

To this end, We have assembled three new datasets from CoSal2015~\cite{zhang2015co}, CoSOD3k~\cite{fan2021re}, and CoCA~\cite{zhang2020gradient}, which we refer to as OWCoSal, OWCoSOD, OWCoCA. Specifically, within each category of each dataset, we introduced noise images in random proportions. We designed two categories of noise images: one where noise images come from the same category and another where noise images come from different categories. This approach allows us to more comprehensively simulate real-world scenarios. The right half of Figure~\ref{fig: dataset} shows some examples of OWCoSal, OWCoSOD, OWCoCA, and the different color border represents the different category.

\paragraph{OWCoSal} 
\begin{figure*}[!t]
\begin{center}
\includegraphics[width=1.0\textwidth]{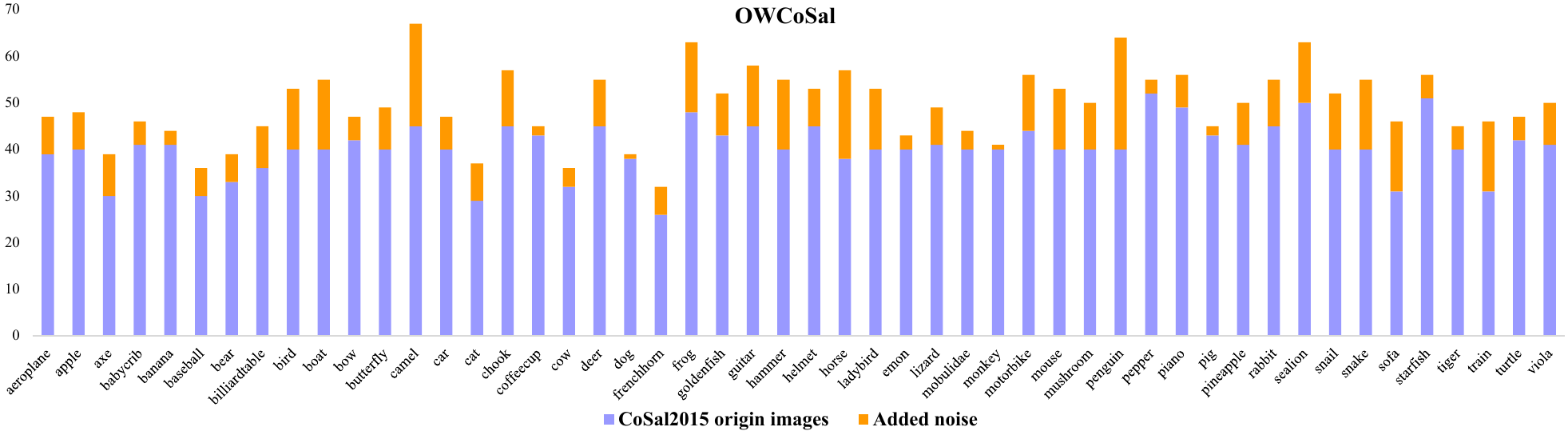}
\end{center}
\caption{The structure of OWCoSal.}
\vspace{-0.4cm}
\label{fig: OWCoSal}
\end{figure*}
As shown in Figure~\ref{fig: OWCoSal}, building upon CoSal2015~\cite{zhang2015co}, we introduced an additional 460 noisy images from different categories, expanding each category by 2.4$\%$ to 37.5$\%$, the majority of them are concentrated around 18$\%$.
In OWCoSal, the added noise images within each category originate from the same category. To control the difficulty of OWCoSal, we kept the number of noise images at a relatively balanced level, around 18$\%$ of the original image count. At the same time, we generated corresponding labels for the noise images, resulting in 460 corresponding all-zero maps.

\paragraph{OWCoSOD} 
\begin{figure*}[!t]
\begin{center}
\includegraphics[width=1.0\textwidth]{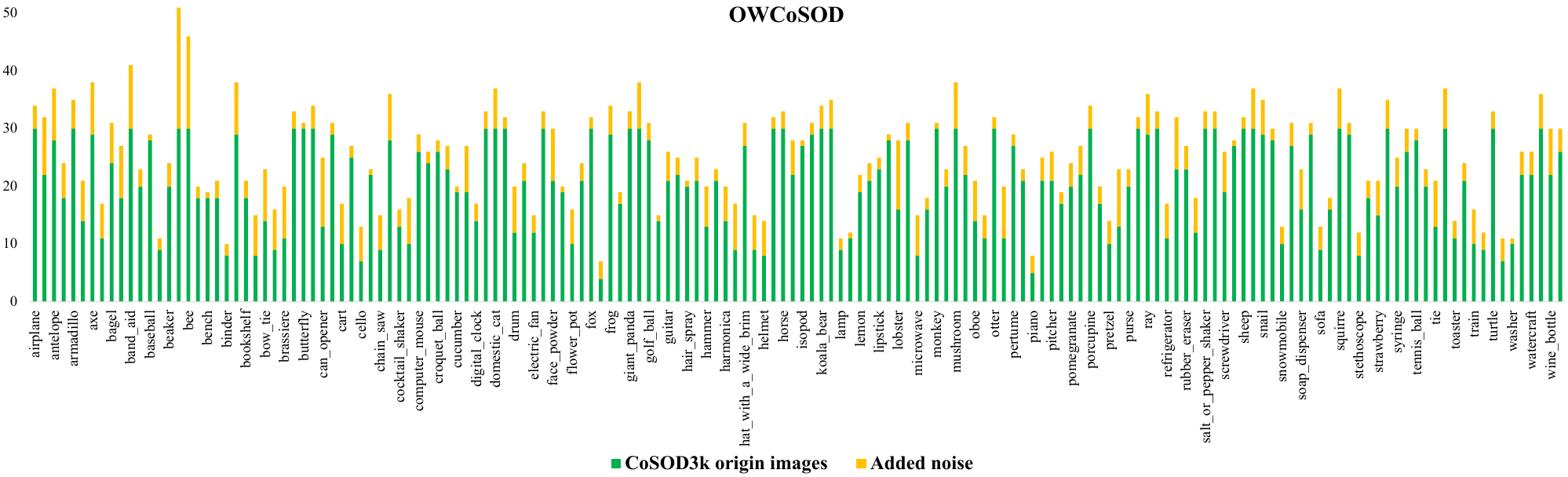}
\end{center}
\caption{The structure of OWCoSal.}
\vspace{-0.4cm}
\label{fig: OWCoSOD}
\end{figure*}
We added 739 noisy images to CoSOD3k~\cite{fan2021re}, following the same approach as in OWCoSal, where the noise added to each category in OWCoSOD also originates from the same category. The expansion ranges from 3.2$\%$ to 47.1$\%$ for different categories. To distinguish it from OWCoSal, the majority of categories in OWCoSOD had expansion percentages concentrated at both ends, at 5$\%$ and 40$\%$. The structure of OWCoSOD is shown in Figure~\ref{fig: OWCoSOD}. The corresponding labels have also been subjected to masking.

\paragraph{OWCoCA} 
\begin{figure*}[!t]
\begin{center}
\includegraphics[width=1.0\textwidth]{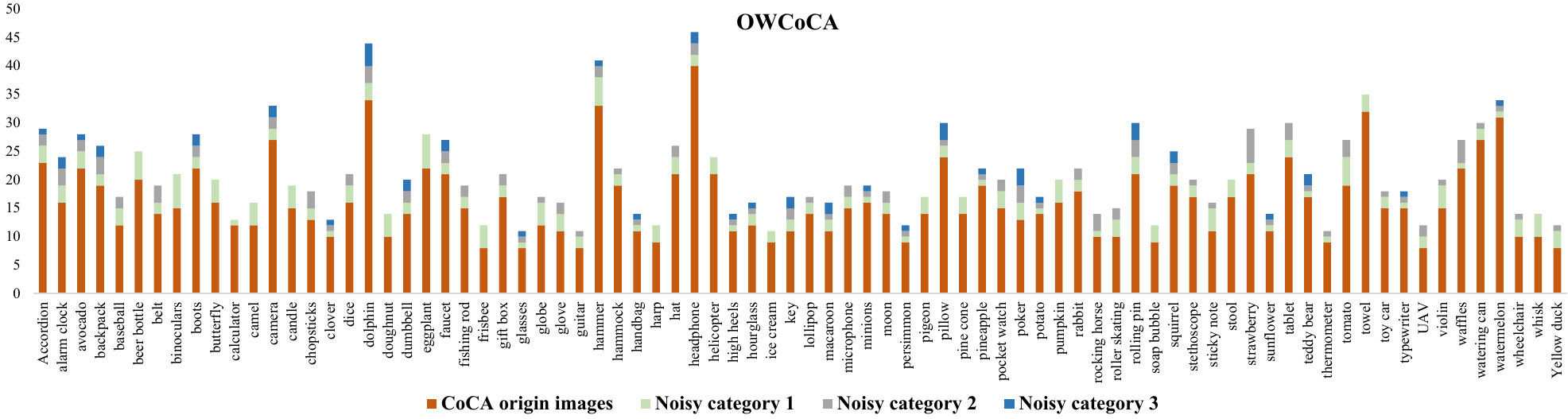}
\end{center}
\caption{The structure of OWCoCA.}
\vspace{-0.4cm}
\label{fig: OWCoCA}
\end{figure*}
As shown in Figure~\ref{fig: OWCoCA}, OWCoCA extends CoCA~\cite{zhang2020gradient} by adding 364 noisy images. Unlike OWCoSal and OWCoSOD, in OWCoCA, the noisy images within each group of images may originate from different categories. We introduced 1-3 categories to simulate open-world scenarios, this makes it the most dynamically changing dataset, and combined with the inherent high difficulty of CoCA, OWCoCA becomes exceedingly challenging. The corresponding labels are generated as OWCoSal and OWCoCA.

\section{Experimental Analysis}\label{sec4}
\subsection{Implementation Details}
We implement our model under the PyTorch framework~\cite{paszke2019pytorch} and the model is accelerated by a single GeForce 3090 GPU. The transformer backbone employed in this work is based on the T2T-ViT$_t$-14 model~\cite{yuan2021tokens} because it exhibits a computational complexity similar to that of the CNN-based ResNet50~\cite{he2016deep} and smaller than VGG-16~\cite{simonyan2014very}, which is commonly used in CoSOD.
In the training process, we used a two-stage training method. In the first stage, we used the COCO-SEG dataset~\cite{wang2019robust} to train the VQ-VAE model to obtain our desired codebook, and on this basis, we continued to train the generative network with the frozen VQ-VAE weights. In the second stage, we used a combination of both COCO-SEG~\cite{wang2019robust} and DUTS datasets~\cite{wang2017learning} to train the entire network. These two datasets contain 208,250 images from 369 categories and the corresponding ground truth for each image. In the second phase of training the whole network, we freeze the model weights of VQ-VAE and only fine-tuned the weights of generative network to ensure codebook quality.

During training, we first randomly select two groups of images and then apply our proposed GSEM strategy with $k=1$ on the two groups to exchange images from each group with each other as noise images. Each training group contains $N=5$ images. The input images are resized to $224\times 224\times 3$. In the first stage, to train the VQ-VAE, we use 4-fold downsampling for $100$ epochs, where the codebook size is set to $128\times 384$, which is represented as $128$ discrete vectors, each with $384$ channels. The generative network is inspired by PixelCNN~\cite{razavi2019generating} and trained with $50$ epochs. The overall training consisted of 60,000 steps, and the Adam optimizer~\cite{kingma2014adam} is used to optimize the whole network. The hyper-parameter in (\ref{eq: bdc}) is set to $\mu=0.5$. The hyper-parameters in (\ref{eq:loss_total}) are set to $\{ \lambda_1=1, \lambda_2=0, \lambda_3=0 \}$, $\{ \lambda_1=0, \lambda_2=1, \lambda_3=0 \}$, $\{ \lambda_1=0, \lambda_2=0, \lambda_3=1 \}$ at different training stage, respectively.
%

%
\subsection{Datasets and Evaluation Metrics}
We utilize the three most widely used benchmark datasets, including CoSal2015\cite{zhang2015co}, CoSOD3k~\cite{fan2020taking}, and CoCA~\cite{zhang2020gradient} to test the models. We employ four commonly used metrics in the CoSOD field to comprehensively evaluate model performance, including $MAE$~\cite{wang2019robust}, $E^{max}_{\phi}$~\cite{fan2018enhanced}, $S_{\alpha}$~\cite{fan2017structure} and $F^{max}_{\beta}$~\cite{achanta2009frequency}.
$MAE$ calculates the average of the absolute differences between each individual observation and its corresponding predicted value. This metric quantifies the average pixel-wise discrepancy between predictions and labels.
$E^{max}_{\phi}$ is an evaluation approach that relies on discrepancies in local pixel-level information as well as global mean information.
$S_m$ assesses the structural similarity with region- and object-awareness similarity between a predicted map and a manually annotated binary label. This metric places greater emphasis on structural information, which quantifies the structural discrepancies between predictions and ground truths.
$\emph{F}_\beta$ is the weighted harmonic mean for recall and precision using non-negative weight.
It quantifies the equilibrium between precision and recall in object retrieval. $\emph{F}_\beta$ is defined as $\frac{(1+\beta^2)Precision \times Recall}{\beta^2 \times Precision + Recall}$, where ${\beta}^2$ is set to 0.3.

\begin{figure*}[!t]
\begin{center}
\includegraphics[width=1.0\textwidth]{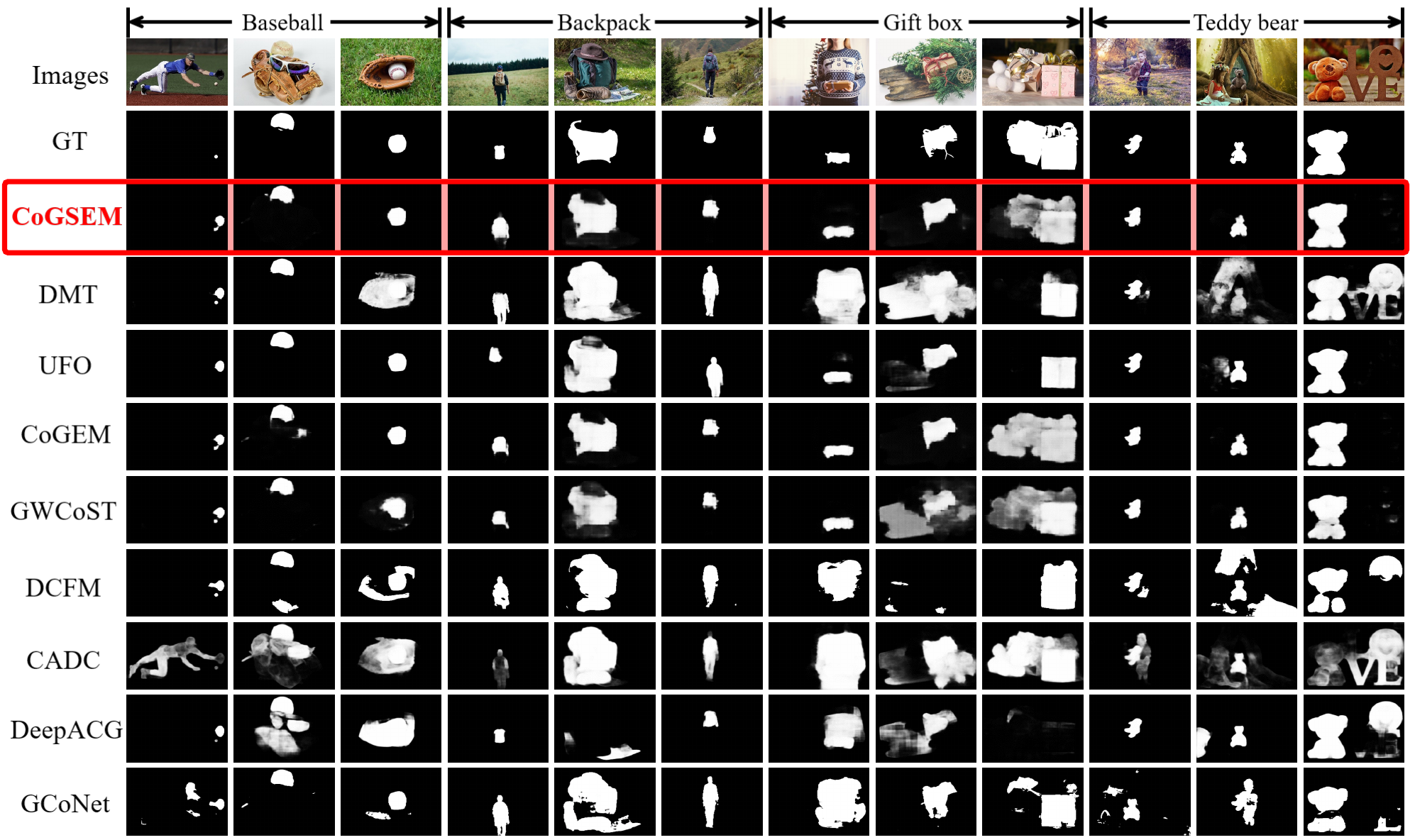}
\end{center}
\caption{Comparison of results with some state-of-the-art methods on the CoCA benchmark, including DMT~\cite{li2023discriminative}, UFO~\cite{su2023unified}, CoGEM~\cite{wu2023co}, GWCoST~\cite{wu2023group}, DCFM~\cite{yu2022democracy}, CADC~\cite{zhang2021summarize}, DeepACG~\cite{zhang2021deepacg}, GCoNet~\cite{fan2021group}.}
\label{fig: comparison}
\end{figure*}

\subsection{Analysis of comparative experimental results}
Using a standardized evaluation tool~\cite{fan2021re}, we compare and analyze our method against those developed in the past five years,
including RCAN~\cite{li2019detecting}, CSMG~\cite{zhang2019co}, SSNM~\cite{zhang2020deep}, GCAGC~\cite{zhang2020adaptive}, GICD~\cite{zhang2020gradient}, ICNet~\cite{jin2020icnet}, GCoNet~\cite{fan2021group}, DeepACG~\cite{zhang2021deepacg}, CoEGNet~\cite{fan2021re}, HrSSMN~\cite{zhang2022deep}, CADC~\cite{zhang2021summarize}, DCFM~\cite{yu2022democracy}, UFO~\cite{su2023unified}, GWCoST~\cite{wu2023group}, CoGEM~\cite{wu2023co}, DMT~\cite{li2023discriminative}, where UFO~\cite{su2023unified}, GWCoST~\cite{wu2023group}, CoGEM~\cite{wu2023co}, DMT~\cite{li2023discriminative} are the latest cutting-edge work.

\paragraph{Qualitative Results.}
We conducted a visual comparison and analysis of our model's predictions against the eight latest cutting-edge methods, including DMT~\cite{li2023discriminative}, UFO~\cite{su2023unified}, CoGEM~\cite{wu2023co}, GWCoST~\cite{wu2023group}, DCFM~\cite{yu2022democracy}, CADC~\cite{zhang2021summarize}, DeepACG~\cite{zhang2021deepacg}, GCoNet~\cite{fan2021group}. The comparative results are shown in Figure~\ref{fig: comparison}.
%
%
The four selected groups are all from CoCA~\cite{zhang2020gradient}, which is the most challenging among the commonly used three benchmark datasets. Choosing more challenging samples better showcases the model's characteristics. These four groups of images are all very difficult, featuring similar interfering objects in terms of shape, extremely complex backgrounds, scenes with camouflage-like properties, significant size variations, and more. These characteristics pose a great test of the model's robustness.
%
%
Specifically, in the group ``Baseball'', there is a significant variation in the appearance of co-salient objects, which challenges the model's ability to grasp co-salient objects, especially when there are larger interfering objects around small objects, models tend to produce incorrect results. DMT, DCFM, CADC, DeepACG, and GCoNet all exhibited varying degrees of misclassification. Additionally, in the second image of this group, the baseball and the background color are very similar, leading to errors in some models, such as DCFM, CADC, DeepACG, and GCoNet.
Benefiting from the introduction of noisy images during training, both CoGSEM and the previous version of CoGEM demonstrated good robustness against interfering objects. Furthermore, due to upgrades in our selection of noisy samples and latent variable representations, CoGSEM outperformed CoGEM in this regard.
In the group ``Backpack'', the difficulty mainly arises from the ``person'' carrying a backpack, which leads to some models incorrectly segmenting the ``person'' as the co-salient target, such as DMT, UFO, and others. The second image in this group is particularly challenging, as the 'backpack' in the image is occluded by interfering objects, and there is a significant size difference compared to the other two images in the same group. DMT, UFO, DCFM, CADC, and GCoNet all incorrectly segmented the interfering objects, while DeepACG even lost the co-salient target. Our CoGSEM demonstrates better robustness and accurately segments the co-salient objects in these challenging scenarios.
The ``Gift box'' group's main characteristic is its camouflage-like nature, which is widely regarded as a challenging scenario in computer vision~\cite{li2021uncertainty,liu2023bi}, and CoSOD is no exception. In the first image of this group, the 'gift box' appears to be part of the sweater pattern, with a very unclear outline. Similarly, the co-salient object in the second image closely resembles the interfering object in terms of color and shape. In the third image, some targets are hidden behind interfering objects. All of these present significant challenges to the model. Other methods exhibited significant errors, but thanks to the fusion of latent variables and the ability to overcome overconfidence, CoGSEM performs well in these scenarios and accurately detects the co-salient objects.
The ``Teddy bear'' group faces similar challenges as the previous three groups, including camouflage-like characteristics, drastic changes in size, and interfering objects with similar colors. Faced with these challenges, other methods tend to make errors in segmenting the background and interfering objects to varying degrees. However, our CoGSEM, building upon the robustness learned from the GSEM strategy and leveraging LVGB and CoSOD-TB, achieves better segmentation accuracy and outperforms other methods when confronted with these challenges.

\begin{figure*}[!t]
\begin{center}
\begin{tabular}{c}
\includegraphics[width=1.0\textwidth]{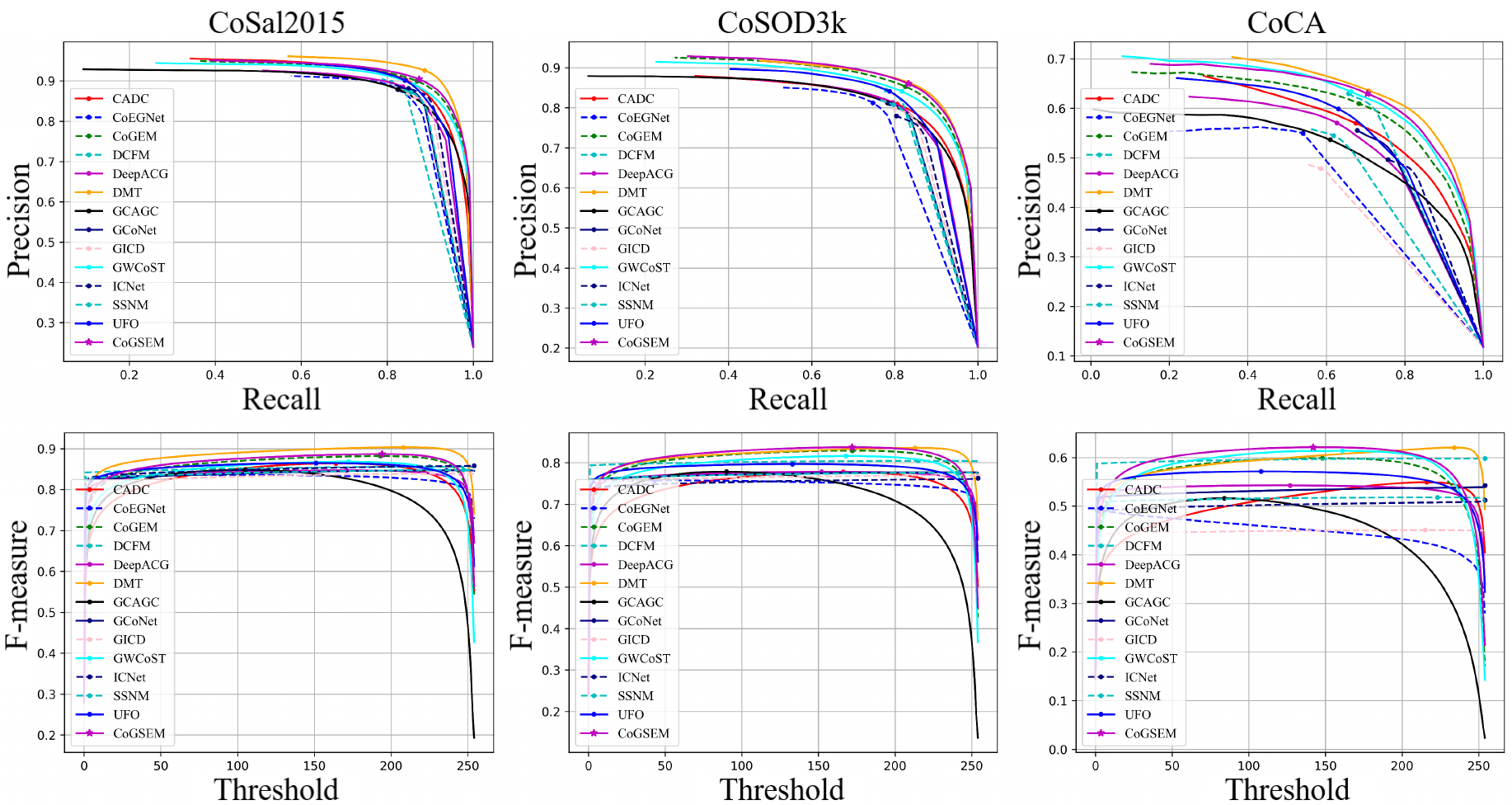}
\end{tabular}
\end{center}
\caption{
Comparison of PR curves and F-measure curves on CoSal2015\cite{zhang2015co}, CoSOD3k~\cite{fan2020taking} and CoCA~\cite{zhang2020gradient} with the leading methods in the past five years.
}
\label{fig: curves}
\end{figure*}
\begin{table*}[!t]
\caption{Statistic comparisons on three benchmark datasets of our model with the other leading methods in the recent five years. \textcolor{red}{Red} represents the best performance, \textcolor{blue}{blue} represents the second-best performance.}
\resizebox{\textwidth}{!}{
\begin{tabular}{c|cccc|cccc|cccc}
\hline
Methods & \multicolumn{4}{c|}{CoSal2015} & \multicolumn{4}{c|}{CoSOD3k}   & \multicolumn{4}{c}{CoCA}      \\
                         & $MAE\downarrow$   & $S_{\alpha}\uparrow$    & $E_{\phi}^{max}\uparrow$    & $F_{\beta}^{max}\uparrow$     & $MAE\downarrow$   & $S_{\alpha}\uparrow$    & $E_{\phi}^{max}\uparrow$    & $F_{\beta}^{max}\uparrow$     & $MAE\downarrow$   & $S_{\alpha}\uparrow$    & $E_{\phi}^{max}\uparrow$    & $F_{\beta}^{max}\uparrow$     \\
                         \hline
RCAN(IJCAI2019)           & 0.126 & 0.779 & 0.842 & 0.764 & 0.130 & 0.744 & 0.808 & 0.688 & 0.160 & 0.616 & 0.702 & 0.422 \\
CSMG(CVPR2019)           & 0.130 & 0.774 & 0.818 & 0.777 & 0.157 & 0.711 & 0.723 & 0.645 & 0.124 & 0.632 & 0.734 & 0.503 \\
SSNM(AAAI2020)           & 0.102 & 0.788 & 0.843 & 0.794 & 0.120 & 0.726 & 0.756 & 0.675 & 0.116 & 0.628 & 0.741 & 0.482 \\
GCAGC(CVPR2020)          & 0.085 & 0.817 & 0.866 & 0.813 & 0.100 & 0.785 & 0.816 & 0.740 & 0.118 & 0.669 & 0.754 & 0.523 \\
GICD(ECCV2020)           & 0.072 & 0.842 & 0.884 & 0.834 & 0.089 & 0.794 & 0.831 & 0.743 & 0.125 & 0.658 & 0.701 & 0.504 \\
ICNet(NIPS2020)          & 0.058 & 0.857 & 0.900 & 0.858 & 0.089 & 0.794 & 0.845 & 0.762 & 0.147 & 0.654 & 0.705 & 0.514 \\
CoEGNet(TPAMI2021)       & 0.077 & 0.836 & 0.882 & 0.832 & 0.092 & 0.762 & 0.825 & 0.736 & 0.106 & 0.612 & 0.717 & 0.493 \\
GCoNet(CVPR2021)         & 0.069 & 0.845 & 0.887 & 0.847 & 0.071 & 0.802 & 0.860 & 0.750 & 0.105 & 0.673 & 0.760 & 0.524 \\
DeepACG(CVPR2021)        & 0.066 & 0.853 & 0.893 & 0.847 & 0.079 & 0.811 & 0.859 & 0.779 & 0.104 & 0.685 & 0.759 & 0.564 \\
CADC(ICCV2021)           & 0.064 & 0.866 & 0.906 & 0.862 & 0.096 & 0.801 & 0.840 & 0.759 & 0.132 & 0.681 & 0.744 & 0.548 \\
HrSSMN(TMM2022)          & 0.062 & 0.845 & 0.895 & 0.841 & 0.087 & 0.788 & 0.842 & 0.753 & 0.106 & 0.671 & 0.739 & 0.532 \\
DCFM(CVPR2022)           & 0.067 & 0.838 & 0.892 & 0.856 & 0.067 & 0.810 & 0.874 & 0.805 & \textcolor{red}{0.085} & 0.710 & 0.783 & 0.598 \\
UFO(TMM2023)             & 0.064 & 0.860 & 0.906 & 0.865 & 0.073 & 0.819 & 0.874 & 0.797 & 0.095 & 0.697 & 0.782 & 0.571 \\
GWCoST(ICASSP2023)       & 0.056 & 0.876 & 0.925 & 0.878 & 0.065 & 0.844 & 0.901 & 0.815 & 0.097 & 0.725 & 0.810 & 0.600 \\
CoGEM(CVPR2023)          & 0.053 & 0.885 & 0.933 & 0.882 & \textcolor{blue}{0.061} & \textcolor{blue}{0.853} & \textcolor{blue}{0.911} & 0.829 & 0.095 & \textcolor{blue}{0.726} & \textcolor{blue}{0.808} & 0.599 \\
DMT(CVPR2023)            & \textcolor{red}{0.047} & \textcolor{red}{0.896} & \textcolor{blue}{0.933} & \textcolor{red}{0.903} & 0.064 & 0.851 & 0.895 & \textcolor{blue}{0.835} & 0.108 & 0.724 & 0.800 & \textcolor{blue}{0.619} \\
CoGSEM(OURS)             & \textcolor{blue}{0.051} & \textcolor{blue}{0.889} & \textcolor{red}{0.936} & \textcolor{blue}{0.887} &  \textcolor{red}{0.059} &  \textcolor{red}{0.860} &  \textcolor{red}{0.915} &  \textcolor{red}{0.837} & \textcolor{blue}{0.092} & \textcolor{red}{0.740} & \textcolor{red}{0.820} & \textcolor{red}{0.621} \\
                         \hline
\end{tabular}}
\label{table: result}
\end{table*}
\paragraph{Quantitative Results.}

In Figure~\ref{fig: curves}, we present the Precision-Recall (PR) curves and F-measure curves of our method and several other methods on three test sets. It can be observed that in the PR curves, our method occupies the outermost or second outermost position. In the F-measure curves, the curves generated by our method also rank the highest or second highest. These all indicate that our CoGSEM has achieved leading performance.

In Table~\ref{table: result}, we conducted a quantitative comparison between CoGSEM and other methods across four metrics on the three benchmark datasets.The results illustrate that our CoGSEM delivers outstanding performance. To be more specific, on the CoSal2015 dataset, our approach attains the highest or the second-highest scores across all metrics, with values of 0.051, 0.889, 0.936, and 0.887. Compared to our previous version CoGEM~\cite{wu2023co}, we achieved improvements of 0.2$\%$, 0.4$\%$, 0.3$\%$, and 0.5$\%$ on the four metrics, respectively.
%
%
%
CoGSEM outperforms other methods on CoSOD3k, surpassing them by 0.059, 0.860, 0.915, and 0.837, across all metrics, respectively. With a gain of 0.5$\%$, 0.9$\%$, 2.0$\%$, and 0.2$\%$ compared to the highly competitive DMT~\cite{li2023discriminative}.
On CoCA, CoGSEM achieved three top rankings and one second ranking among the four metrics, with scores of 0.092, 0.740, 0.820, and 0.621, respectively. Among them, $S_{\alpha}$ and $E_{\phi}^{max}$  achieve 1.4$\%$ and 1.2$\%$ improvements compared to the second-best method, respectively
CoSOD3k and CoCA are the latest and most challenging publicly available test datasets. The higher level of performance achieved on CoSOD3k and CoCA indicates that our method excels in demonstrating outstanding robustness and competitiveness when faced with more diverse and complex scenarios.

\begin{table*}[!t]
\caption{Ablation experiments to assess the effectiveness of our designed GSEM, LVGB, CoSOD-TB, LVGB$_p$, and GEM and LVGB in our method. \textcolor{red}{Red} represents the best performance, \textcolor{blue}{blue} represents the second-best performance.}
\resizebox{\textwidth}{!}
{
    \begin{tabular}{ccccc|cccc|cccc|cccc}
    \hline
    \multicolumn{5}{c|}{Strategies}                     & \multicolumn{4}{c|}{CoSal2015} & \multicolumn{4}{c|}{CoSOD3k}                             & \multicolumn{4}{c}{CoCA}                                        \\
    GSEM & LVGB & CoSOD-TB & LVGB$_p$ & GEM & $MAE\downarrow$  & $S_{\alpha}\uparrow$ & $E_{\phi}^{max}\uparrow$ & $F_{\beta}^{max}\uparrow$ & \multicolumn{1}{c}{$MAE\downarrow$} & \multicolumn{1}{c}{$S_{\alpha}\uparrow$} & $E_{\phi}^{max}\uparrow$ & $F_{\beta}^{max}\uparrow$ & \multicolumn{1}{c}{$MAE\downarrow$} & \multicolumn{1}{c}{$S_{\alpha}\uparrow$} & $E_{\phi}^{max}\uparrow$     & $F_{\beta}^{max}\uparrow$     \\
    \hline
     &  &  & & & 0.060 & 0.854 & 0.887 & 0.860 & 0.075 & 0.787 & 0.863 & 0.778 & 0.105 & 0.710 & 0.785 & 0.564 \\
     &  &  & & $\checkmark$ & 0.061 & 0.883 & 0.928 & 0.877 & 0.065 & 0.834 & 0.880 & 0.817 & 0.109 & 0.710 & 0.788 & 0.571 \\
     $\checkmark$ &  &  & & & 0.063 & 0.890 & 0.925 & 0.873 & 0.064 & 0.836 & 0.888 & 0.824 & 0.106 & 0.717 & 0.792 & 0.578 \\
     &  &  & $\checkmark$ & & 0.058 & 0.877 & 0.926 & 0.872 & 0.063 & 0.840 & 0.871 & 0.805 & 0.100 & 0.719 & 0.789 & 0.574 \\
     & $\checkmark$ &  & & & 0.056 & 0.882 & 0.930 & 0.875 & 0.063 & 0.838 & 0.879 & 0.812 & 0.099 & 0.724 & 0.794 & 0.579 \\
     &  & $\checkmark$ & & & 0.053 & 0.874 & 0.893 & 0.880 & 0.062 & 0.842 & 0.900 & 0.814 & 0.098 & 0.716 & 0.798 & 0.583 \\
     &  & $\checkmark$ & $\checkmark$ & & 0.055 & 0.869 & 0.930 & 0.876 & 0.063 & 0.847 & 0.906 & 0.819 & 0.104 & 0.724 & 0.802 & 0.597 \\
     & $\checkmark$ & $\checkmark$ &  & & 0.053 & 0.877 & 0.932 & \textcolor{blue}{0.885} & 0.065 & 0.847 & 0.915 & 0.820 & 0.104 & \textcolor{blue}{0.736} & 0.807 & 0.603 \\
     &  &  & $\checkmark$ & $\checkmark$ & 0.060 & \textcolor{blue}{0.886} & 0.925 & 0.872 & 0.069 & 0.850 & 0.895 & 0.823 & 0.100 & 0.718  & 0.792 & 0.587 \\
     $\checkmark$& $\checkmark$ &  &  &  & 0.062 & 0.883 & 0.928 & 0.872 & 0.067 & \textcolor{blue}{0.856} & 0.904 & 0.831 & 0.098 & 0.721  & 0.801 & 0.594 \\
     &  & $\checkmark$ & & $\checkmark$ & 0.054 & 0.880 & 0.919 & 0.878 & 0.061 & 0.849 & 0.889 & 0.829 & 0.096 & 0.720  & 0.805 & 0.595 \\
     $\checkmark$ &  & $\checkmark$ & & & \textcolor{blue}{0.052} & 0.876 & 0.922 & 0.882 & \textcolor{blue}{0.058} & 0.855 & 0.901 & \textcolor{blue}{0.834} & \textcolor{blue}{0.094} & 0.731  & \textcolor{blue}{0.812} & \textcolor{blue}{0.614} \\
     &  & $\checkmark$ & $\checkmark$ & $\checkmark$ & 0.053 & 0.885 & \textcolor{blue}{0.933} & 0.882 & 0.061 & 0.853 & \textcolor{blue}{0.911} & 0.829 & 0.095 & 0.726 & 0.808 & 0.599  \\
     $\checkmark$ & $\checkmark$ & $\checkmark$ &  &  & \textcolor{red}{0.051} & \textcolor{red}{0.889} & \textcolor{red}{0.936} & \textcolor{red}{0.887} &  \textcolor{red}{0.059} &  \textcolor{red}{0.860} &  \textcolor{red}{0.915} &  \textcolor{red}{0.837} & \textcolor{red}{0.092} & \textcolor{red}{0.740} & \textcolor{red}{0.820} & \textcolor{red}{0.621}  \\
    \hline
    \end{tabular}
}
\vspace{-0.4cm}
\label{table: ablation}
\end{table*}
\subsection{Ablation Study}
To validate the effectiveness of the key designs in our CoGSEM, and the improvements over the previous version CoGEM~\cite{wu2023co}, we conducted ablation experiments on three datasets. Our baseline method is Visual saliency transformer (VST)~\cite{liu2021visual}, it is the first to propose a framework entirely based on transformers in the field of saliency detection, which is both concise and efficient, and has achieved outstanding results. Building upon this foundation, we have made modifications and introduced new designs
From Table~\ref{table: ablation}, we can observe that our design aligns well with the characteristics of CoSOD tasks, and each key design contributes to the improvement in model performance.
Taking the results in CoCA as examples, when not using GSEM, there was a noticeable decline in model performance, with all four metrics showing a decrease. Specifically, $E_{\phi}^{max}$ and $F_{\beta}^{max}$ decreased by 1.3$\%$ and 1.8$\%$, respectively. This is because our designed GSEM effectively enhances the model's robustness through noise-introduced training. We also compared GSEM with the previous version GEM, when using only GSEM or GSM, their differences in the four metrics are 0.3$\%$, 0.7$\%$, 0.4$\%$, and 0.7$\%$, respectively. We can observe that GSEM is more effective.
Without LVGB, all four metrics were negatively affected. $MAE$ increased from 0.092 to 0.094, which is a 0.2$\%$ deterioration. S decreased by 0.9$\%$, going from 0.740 to 0.731, $E_{\phi}^{max}$ decreased by 0.8$\%$, going from 0.820 to 0.812, and $F_{\beta}^{max}$ also decreased by 0.7$\%$, dropping from 0.621 to 0.614. This indicates that our designed LVGB helps the model to consider more possibilities from the input images, avoiding overconfidence in the early stages and improving the localization and segmentation of co-salient objects. Similarly, we compared LVGB with the previous version LVGB$_P$. When using only LVGB or LVGB$_P$, LVGB performs better in extracting components with stochastic characteristics, resulting in improved model performance. It outperforms LVGB$_P$ by 0.1$\%$, 0.5$\%$, 1.0$\%$, and 0.5$\%$ on the four metrics, respectively.
Our designed CoSOD-TB is also a crucial component, as it models long-range dependencies in a sequence-to-sequence manner, enabling precise segmentation of object details. When applied in conjunction with GSEM and LVGB, the use of CoSOD-TB further enhances model performance, resulting in improvements of 0.6$\%$, 1.9$\%$, 1.9$\%$, and 2.7$\%$ on the four metrics, respectively. 
It should be noted that applying any of our individual designs in isolation may not necessarily yield favorable results in some cases. For instance, when using GSEM alone, the $MAE$ metric may decrease. This is because the introduction of noise can interfere with the common features among images in the same group, further affecting the model's ability to capture the details of each image.

\subsection{Practical Application}
\begin{figure*}[!t]
\begin{center}
\includegraphics[width=1\textwidth]{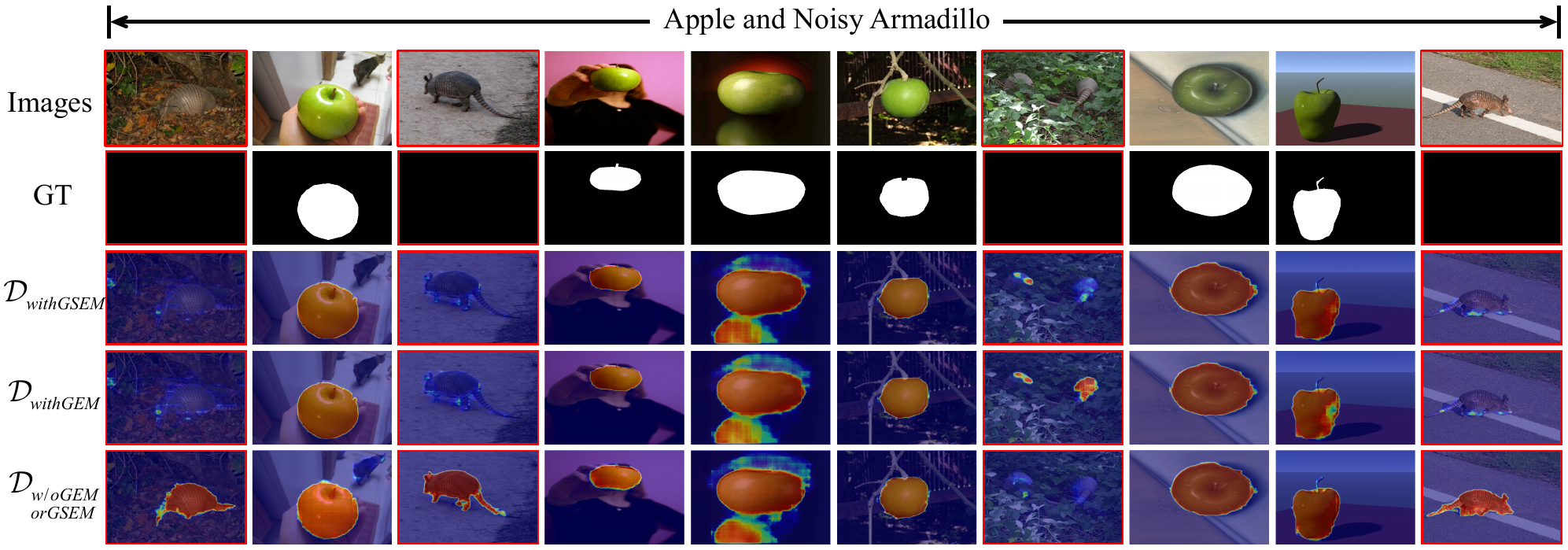}
\end{center}
\caption{The response maps with either GSEM or GEM or neither GSEM nor GEM. The first row shows an image group with noisy samples that come from another category. The second row is the ground truth. The third and fourth rows show the response maps $\mathcal{D}_{withGEM}$ and $\mathcal{D}_{w/oGEM}$ from the decoder.}
\label{fig: gsemvisual}
\end{figure*}
\begin{figure*}[!t]
\begin{center}
\includegraphics[width=0.5\textwidth]{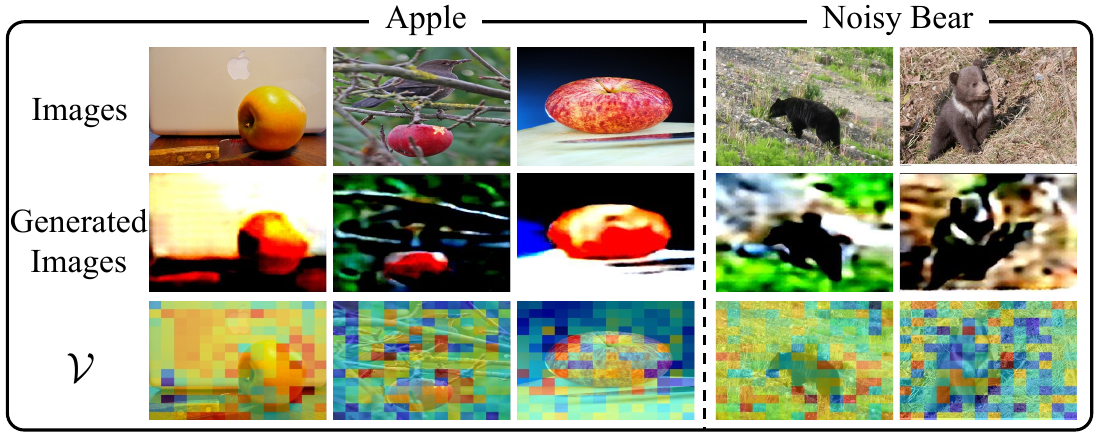}
\end{center}
\caption{Image generation results and intermediate feature visualization of VQ-VAE.}
\label{fig: vaevisual}
\end{figure*}
To better assess the model's detection and segmentation capabilities in open-world scenarios, we conducted further experiments in simulated real-world environments and evaluated the model's performance on the three open-world datasets introduced in this paper: OWCoSal, OWCoSOD, and OWCoCA.
We conducted a comparison of feature visualization results for noisy groups. As shown in Figure~\ref{fig: gsemvisual}, we provide visual results of CoGSEM during the sampling stage of the model when ``Armadillo'' noise is introduced in the ``apple'' group. It can be observed that the use of GSEM effectively enhances the model's robustness, resulting in superior performance compared to GEM. This allows the model to better suppress noisy images. However, when GSEM or GEM is not used, the model's robustness significantly decreases, leading to incorrect segmentation of non-co-salient objects in noisy images. This is detrimental to the model's applicability in open-world scenarios.
Figure~\ref{fig: vaevisual} displays the image generation results in LVGB along with their corresponding feature heatmaps in the decoder. As we can observe, our designed LVGB can implicitly model group global features while accommodating the generation of stochastic features $\mathcal{V}$. For the primary target ``Apple'', its heat map focuses more on the target region. For the noise ``Bear'', its heat map is more dispersed, resulting in a blurrier generation of the objects in the image. In the subsequent fusion process, the low-contrast heat maps of noise images are effective in suppressing the noise images, while the heat maps of images containing co-salient objects are useful in overcoming overconfidence.

\begin{figure*}[!t]
\begin{center}
\includegraphics[width=1\textwidth]{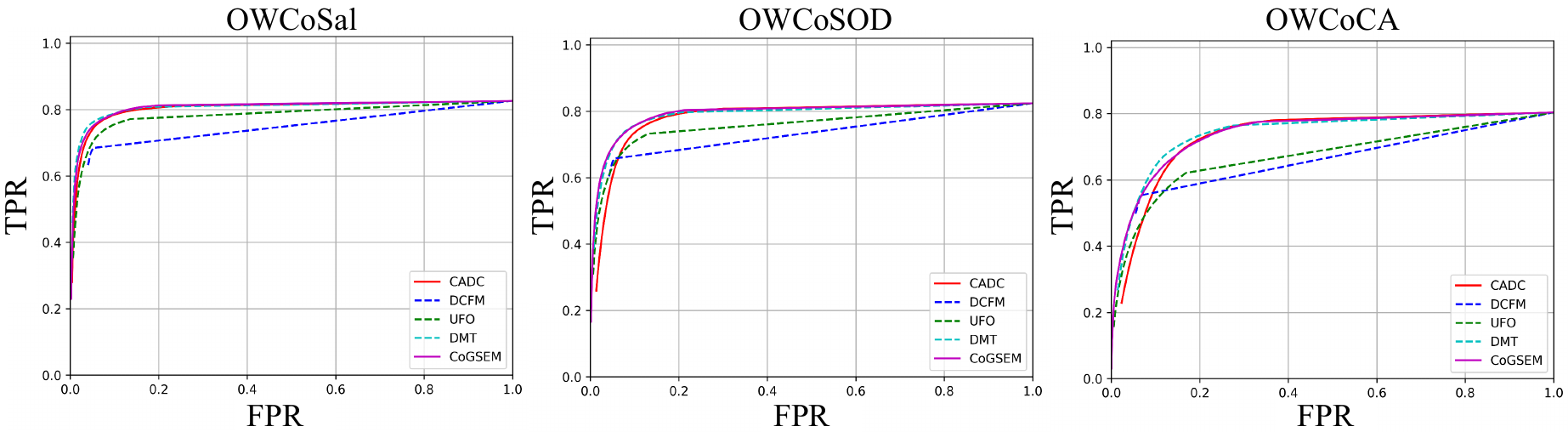}
\end{center}
\caption{ROC of CoGSEM and several state-of-the-arts.}
\label{fig: roc}
\end{figure*}

%
We also conducted some quantitative analyses. Table~\ref{table: OWresult} provides detailed test data on the open-world datasets. The introduction of noisy images has significantly increased the difficulty of these three datasets. Compared to their performance on datasets CoSal2015~\cite{zhang2015co}, CoSOD3k~\cite{fan2021re} and CoCA~\cite{zhang2020gradient}, previously representative methods DMT~\cite{li2023discriminative}, UFO~\cite{su2023unified}, DCFM~\cite{yu2022democracy} and CADC~\cite{zhang2021summarize} have all shown a notable decline in performance. In contrast, the performance of CoGSEM does not exhibit a significant decline, and in some cases, even showed improvement in certain metrics. 
As we can see, our method achieves the best results on eight out of twelve metrics, while ranking second on the remaining three. CoGSEM demonstrated a significant advantage in terms of the $MAE$ and $S_{\alpha}$ metrics, outperforming DMT~\cite{li2023discriminative} by 2.4$\%$ on the most challenging dataset OWCoCA, outperforming UFO~\cite{su2023unified} by 2.6$\%$ on the largest dataset OWCoSOD, respectively.
Figure~\ref{fig: roc} displays the receiver operating characteristic curves (ROC), and from the graph, it can be observed that CoGSEM is positioned in the top-left corner, maintaining a lower false positive rate even with the highest correct detection, once again affirming CoGSEM's superior performance.

\begin{figure*}[!t]
\begin{center}
\includegraphics[width=1.0\textwidth]{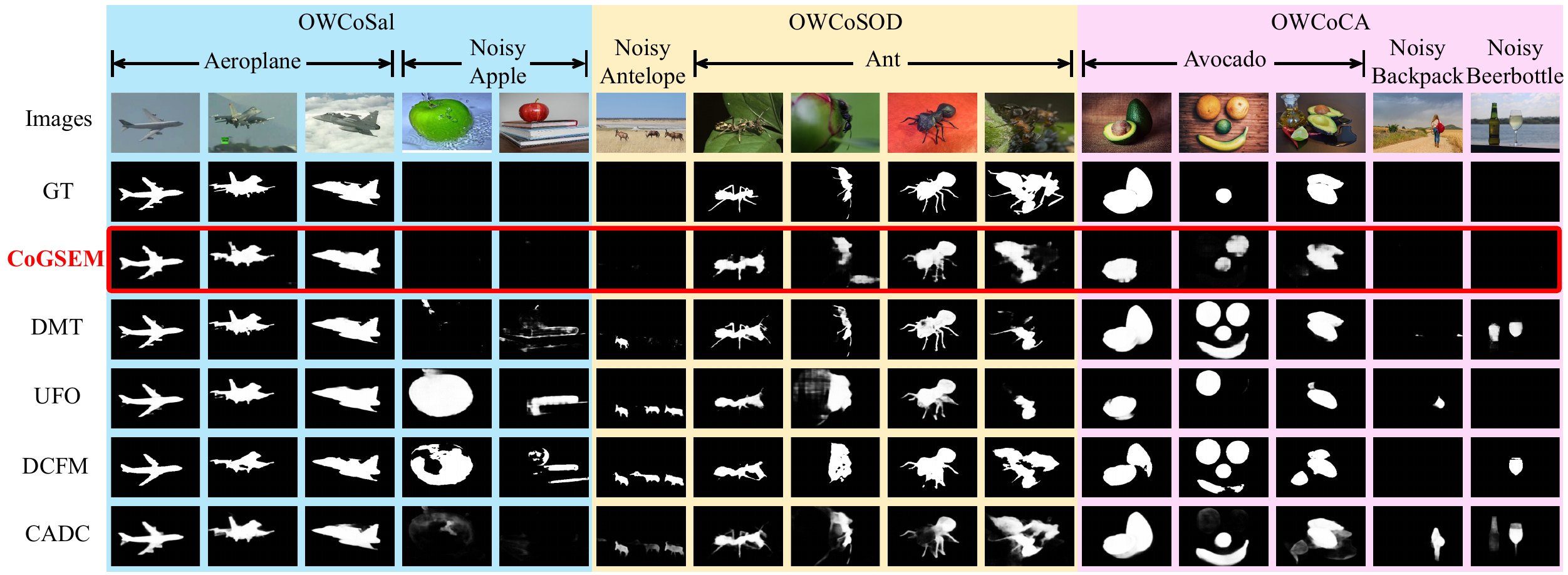}
\end{center}
\caption{The qualitative comparisons on open-world benchmark datasets with the recent state-of-the-art methods, including DMT~\cite{li2023discriminative}, UFO~\cite{su2023unified}, DCFM~\cite{yu2022democracy}, CADC~\cite{zhang2021summarize}, }
\label{fig: OWcomparison}
\end{figure*}

At last, we also proceed to visually compare the results of these methods on the three open-world test datasets. From Figure~\ref{fig: OWcomparison}, we can observe that the introduction of noise is a challenge for all the models. 
Among these datasets, OWCoSal includes a higher proportion of noisy images, OWCoSOD contains a lower proportion of noisy images, and OWCoCA includes noisy images from different categories. Our CoGSEM method demonstrates robustness not only against noisy images but also in achieving more accurate co-salient object segmentation.
Specifically, taking OWCoCA as an example, due to the strong interference from noisy images within the image groups, other methods all exhibit wrong segmentation of co-salient objects. Moreover, these methods display high confidence in their erroneous segmentation. DMT even goes so far as to segment all salient objects in the second image of ``Avocado'', disregarding the crucial attribute of ``co-saliency'' in the CoSOD task.
Furthermore, DMT, UFO, and DCFM all make incorrect segmentation of noisy images, and CADC exhibits poorer robustness when dealing with noisy images, segmenting the most noisy regions. In contrast, our approach, CoGSEM, benefits from the GSEM strategy and LVGB, demonstrating strong robustness to noisy images. It effectively handles challenging samples like the second image of ``Avocado'', correctly segments co-salient objects, and assigns lower confidence to interference objects that were wrong segmented. This further underscores the superiority of our method.

\begin{table*}[!t]
\caption{Statistic comparisons of our model with CADC~\cite{zhang2021summarize}, DCFM~\cite{yu2022democracy}, DMT~\cite{li2023discriminative}, UFO~\cite{su2023unified} on three open-world benchmark datasets. \textcolor{red}{Red} represents the best performance, \textcolor{blue}{blue} represents the second-best performance.}
\resizebox{\textwidth}{!}{
\begin{tabular}{c|cccc|cccc|cccc}
\hline
Methods & \multicolumn{4}{c|}{OWCoSal} & \multicolumn{4}{c|}{OWCoSOD}   & \multicolumn{4}{c}{OWCoCA}      \\
                         & $MAE\downarrow$   & $S_{\alpha}\uparrow$    & $E_{\phi}^{max}\uparrow$    & $F_{\beta}^{max}\uparrow$     & $MAE\downarrow$   & $S_{\alpha}\uparrow$    & $E_{\phi}^{max}\uparrow$    & $F_{\beta}^{max}\uparrow$     & $MAE\downarrow$   & $S_{\alpha}\uparrow$    & $E_{\phi}^{max}\uparrow$    & $F_{\beta}^{max}\uparrow$     \\
                         \hline
CADC(ICCV2021)           & 0.068 & 0.869 & 0.780 & 0.699 & 0.096 & 0.818 & 0.741 & 0.630 & 0.136 & 0.707 & 0.639 & 0.425 \\
DCFM(CVPR2022)           & 0.084 & 0.834 & 0.770 & 0.695 & 0.081 & 0.815 & 0.760 & 0.657 & \textcolor{blue}{0.087} & 0.742 & 0.666 & 0.468 \\
UFO(TMM2023)             & 0.076 & 0.853 & 0.780 & 0.695 & 0.082 & 0.821 & 0.753 & 0.637 & 0.095 & 0.720 & 0.649 & 0.419 \\
DMT(CVPR2023)            & \textcolor{blue}{0.057} & \textcolor{red}{0.890} & \textcolor{blue}{0.804} & \textcolor{red}{0.733} & \textcolor{blue}{0.078} & \textcolor{blue}{0.848} & \textcolor{blue}{0.779} & \textcolor{red}{0.682} & 0.114 & \textcolor{blue}{0.745} & \textcolor{blue}{0.679} & \textcolor{red}{0.479} \\
CoGSEM(OURS)             & \textcolor{red}{0.053} & \textcolor{blue}{0.887} & \textcolor{red}{0.805} & \textcolor{blue}{0.714} &  \textcolor{red}{0.068} &  \textcolor{red}{0.850} &  \textcolor{red}{0.780} &  \textcolor{blue}{0.663} & \textcolor{red}{0.090} & \textcolor{red}{0.754} & \textcolor{red}{0.686} & \textcolor{blue}{0.469} \\
                         \hline
\end{tabular}}
\label{table: OWresult}
\end{table*}
\begin{figure*}[!t]
\begin{center}
\includegraphics[width=0.5\textwidth]{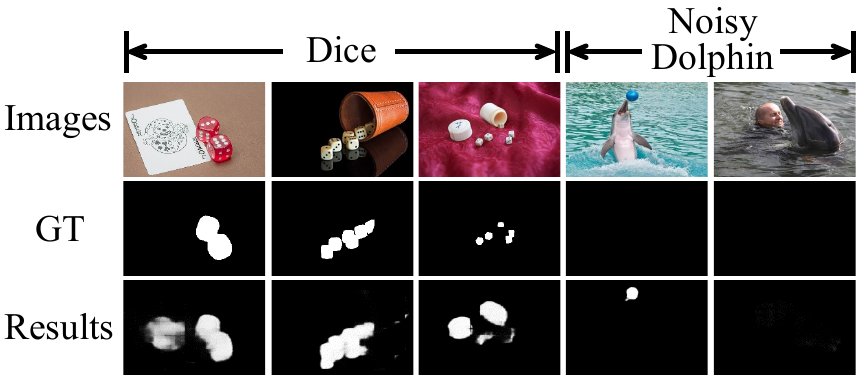}
\end{center}
\caption{Failure cases of our method sampled from OWCoCA.}
\label{fig: faliurecase}
\end{figure*}

\subsection{Failure Cases}
To provide a more comprehensive analysis of our model's performance, we also present some failure cases. From Figure~\ref{fig: faliurecase}, it can be observed that our model performs poorly when faced with extremely challenging samples, including small objects, ghosting, and the presence of similar interfering objects. When similar-looking objects appear in the noisy images, the model also exhibits a certain degree of wrong detection.

\section{Conclusion}\label{sec5}
In this paper, we have extensively and thoroughly discussed the group consensus assumption in CoSOD. We have identified the shortcomings of this assumption, particularly in scenarios where image groups collected may include noisy images. This assumption significantly impacts the model's robustness, thereby affecting its applicability in open-world settings.
We have introduced a learning framework that is distinct from previous classic methods.
%
First, we have enhanced the group exchange-masking strategy to become group-selective exchange-masking.GSEM comprehensively assesses the difficulty of images through a blended measure. With GSEM, we can select the most challenging images from two groups and swap them between the groups to facilitate the model in acquiring more robust representations.
%
Second, we have introduced a latent variable generator based on VQ-VAE, which extracts discrete latent variables to better represent stochastic features. This helps other branches' features overcome overconfidence and focus more effectively on co-salient objects. 
Third, we developed the CoSOD transformer branch to capture global characteristics based on correlations, which contain information regarding group consistency. The features from these branches are concatenated and passed through a transformer-based decoder, resulting in the generation of high-quality co-saliency maps.
Fourth, we introduced three datasets tailored for open-world scenarios, namely, OWCoSal, OWCoSOD, and OWCoCA. These datasets effectively evaluate the model's performance in real-world applications, thus contributing to the advancement of CoSOD field.
Extensive evaluations on three of the most commonly used benchmark datasets and the newly proposed three open-world benchmark test sets have demonstrated the superiority of our method.

\backmatter










\bibliography{sn-bibliography}

\end{document}